\documentclass[times,twocolumn,final,authoryear]{elsarticle}

\usepackage{jasr}
\usepackage{framed,multirow}
\usepackage{latexsym}
\usepackage{url}
\usepackage{hyperref}

\journal{Journal of Advances in Space Research}
\usepackage{forest}

\usepackage{grffile}
\usepackage{alltt}
\usepackage{booktabs}
\usepackage{graphicx}
\usepackage{multicol}
\usepackage{amsmath,amssymb,amsfonts}
\usepackage{algorithmic}
\usepackage{algorithm}
\usepackage{mathtools}
\usepackage{amsthm}
\usepackage{amsmath}
\usepackage{amssymb}
\usepackage{units}
\usepackage{xcolor}
\definecolor{newcolor}{rgb}{.8,.349,.1}
\usepackage{subfigure}
\DeclareGraphicsExtensions{.pdf,.png}

\usepackage[]{units}
\usepackage{setspace}
\usepackage{float}
\def\BibTeX{{\rm B\kern-.05em{\sc i\kern-.025em b}\kern-.08em
    T\kern-.1667em\lower.7ex\hbox{E}\kern-.125emX}}
\newcommand*{\smallmat}[1]
  {\left[\begin{smallmatrix}#1\end{smallmatrix}\right]}

\begin{document}

\verso{Akash Patel \textit{etal}}

\begin{frontmatter}

\title{Towards Energy Efficient Autonomous Exploration of Mars Lava Tube with a Martian Coaxial Quadrotor}

\author[1]{Akash \snm{Patel}\corref{cor1}}
\cortext[cor1]{Corresponding author}
\ead{akash.patel@ltu.se}

\author[1]{Samuel \snm{Karlsson}}
\author[1]{Bj\"orn \snm{Lindqvist}}
\author[1]{Christoforos \snm{Kanellakis}}
\author[]{Ali-Akbar \snm{Agha-Mohammadi}$^4$}

\author[1]{George \snm{Nikolakopoulos}}

\fntext[fn1]{This work has been partially funded by the European Unions Horizon 2020 Research and Innovation Programme under the Grant Agreement No. 869379 illuMINEation.}
\fntext[fn2]{The video link of the presented work: \url{https://youtu.be/9mc0UzAJC3U}}
\address[1]{Robotics \& AI Team, Department of Computer, Electrical and Space Engineering,\\ Lule\r{a} University of Technology, Lule\r{a} SE-97187, Sweden}
\fntext[fn3]{This work was carried out at Lule\r{a} University of Technology, Lulea, Sweden and not in the author's capacity as an employee of JPL, California Institute of Technology.}

\fntext[fn4]{The author is an employee of Jet Propulsion Laboratory, California Institute of Technology Pasadena, CA, 91109, USA}

\received{}
\finalform{}
\accepted{}
\availableonline{}
\communicated{}
\begin{abstract}
Mapping and exploration of a Martian terrain with an aerial vehicle has become an emerging research direction, since the successful flight demonstration of the Mars helicopter Ingenuity. Although the autonomy and navigation capability of the state of the art Mars helicopter has proven to be efficient in an open environment, the next area of interest for exploration on Mars are caves or ancient lava tube like environments, especially towards the never-ending search of life on other planets. This article presents an autonomous exploration mission based on a modified frontier approach along with a risk aware planning and integrated collision avoidance scheme with a special focus on energy aspects of a custom designed Mars Coaxial Quadrotor (MCQ) in a Martian simulated lava tube. One of the biggest novelties of the article stems from addressing the exploration capability, while rapidly exploring in local areas and intelligently global re-positioning of the MCQ when reaching dead ends in order to to efficiently use the battery based consumed energy, while increasing the volume of the exploration. The proposed novel algorithm for the Martian exploration is able to select the next way point of interest, such that the MCQ keeps its heading towards the local exploration direction where it will find maximum information about the surroundings. The proposed three layer cost based global re-position point selection assists in rapidly redirecting the MCQ to previously partially seen areas that could lead to more unexplored part of the lava tube. The Martian fully simulated mission presented in this article takes into consideration the fidelity of physics of Mars condition in terms of thin atmosphere, low surface pressure and low gravity of the planet, while proves the efficiency of the proposed scheme in exploring an area that is particularly challenging due to the subterranean-like environment. The proposed exploration-planning framework is also validated in simulation by comparing it against the graph based exploration planner. Intensive simulations with true Mars conditions are carried out in order to validate and benchmark our approach in a utmost realistic Mars lava tube exploration scenario using a Mars Coaxial Quadrotor.
\end{abstract}


\end{frontmatter}


\section{Introduction}\label{sec:introduction}

For many years, Mars has been a prime candidate for terrestrial exploration in the solar system. The findings of Mars exploration lenders and rovers have paved a way for the future exploration of the planet with much ambitious missions that also include aerial surveying and mapping of the terrain with a helicopter and with increased levels of autonomy. The Mars exploration rovers have presented limitations in terms of their pace, traversability, as well as exploration capability from the ground. To address some of these limitations and mainly for the technology demonstration purposes, a Mars helicopter named Ingenuity~\citep*{balaram2018mars} was sent to Mars with the Perseverance rover~\citep*{farley2020mars}. Recent observations~\citep*{cushing2007themis,cushing2012candidate} from operational satellites in Mars orbit have shown evidence of possible cave skylights, subsurface voids and potential lava tubes like structures on Mars. Exploration of such environment is particularly interesting because upon surveying the environment with a robot, the structural integrity can be inspected for possible habitation inside a cave or void. Traversing through a cave like environment poses multiple challenges to wheel based rovers due to the uneven ground, rocks and boulders. MAVs are particularly interesting platforms because of the increased degrees of freedom, while providing a faster exploration independent of the terrain but limited in time. Due to the reason that very little or no information about the surrounding is available at the start of the exploration, it is necessary that the MCQ is backed by a sophisticated on board autonomy for guidance, navigation and control (GNC).

\begin{figure*}[h!]
    \centering
        \includegraphics[width=\linewidth]{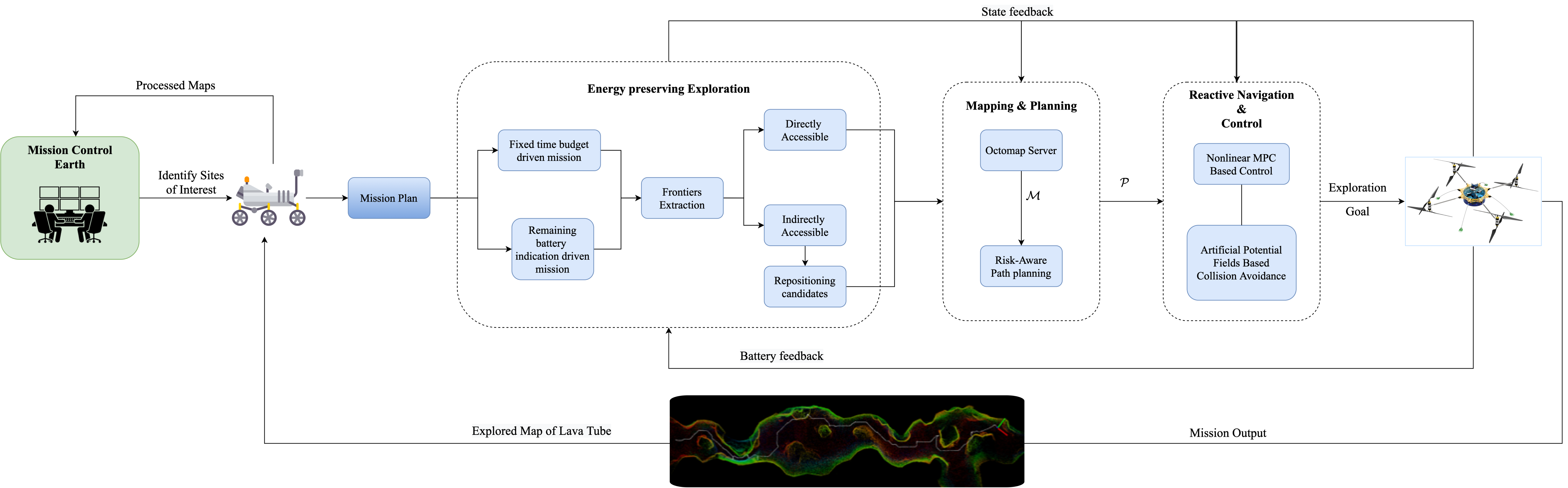}
    {\caption{Mars Lava Tube exploration mission configuration and overall MCQ autonomy architecture}}
    \label{fig:graphicalabstract}
\end{figure*}

Among the essential components of such GNC architecture is the exploration algorithm that feeds in the information about where the robot should navigate next. Exploration task also involves incrementally building a map of the environment as the robot navigates further. Occupancy grids composed of Voxels or cells are used to represent the environment around the robot. Each cell has a value of occupancy probability which represents if the cell is free, occupied or unknown. Based on the occupancy probability of a particular cell is defined to be frontier if it falls within the bounds set by the algorithm. The exploration algorithm also iterates over frontiers which selects an optimal frontier cell out of the list of frontiers. The optimal frontier is selected such that if a robot navigates to that point, it will acquire maximum information about the surrounding environment to extend the map. The method by which an occupancy probability is assigned mainly depends on the sensor used to map the area. Mainly these sensor feed point cloud data to the frontier based exploration algorithm and an occupancy probability values is assigned to each cell. This article presents a fully realistic simulated mission at a Martian atmosphere and environment, in which a coaxial Quadrotor, designed specifically to operate in Mars condition, explores a cave or lava tube like environment consisting of multiple junctions, wide as well as confined passages and unstructured interior that closely resembles an actual Martian cave. The overall mission architecture with proposed autonomy workflow is presented in~\autoref{fig:graphicalabstract}.

\begin{figure}[h!]
    \centering
    \subfigure[]
    {
        \includegraphics[width=0.47\linewidth]{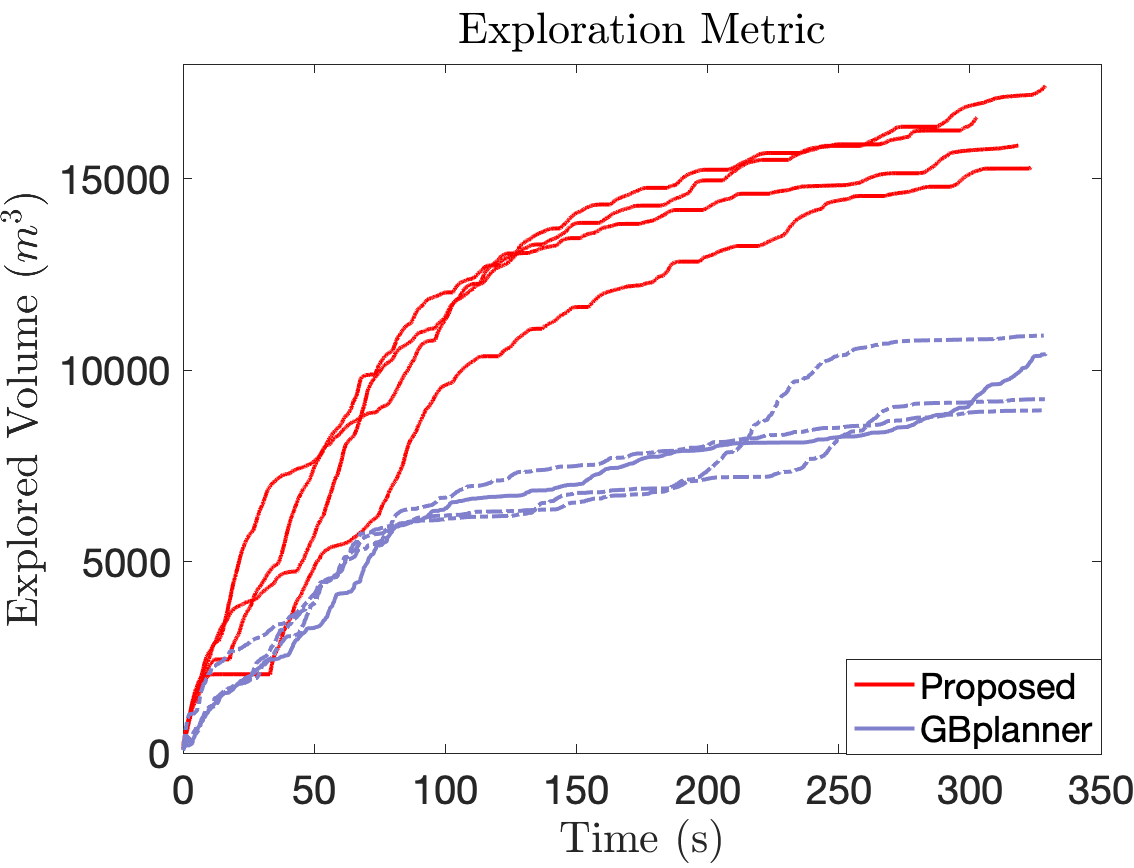}
        
    }
    \subfigure[]
    {
        \includegraphics[width=0.47\linewidth]{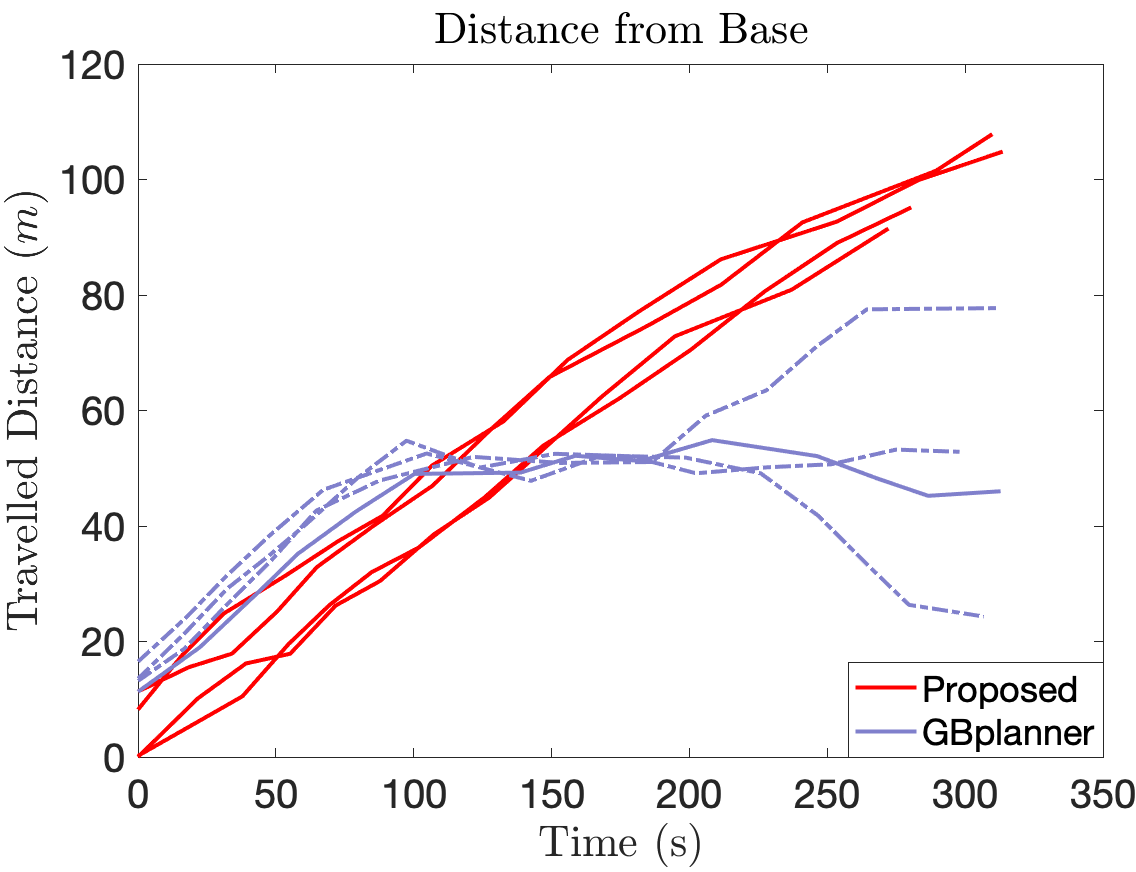}
        
    }
    {\caption{Exploration metrics comparison with state of the art global exploration planner.}}
    \label{fig:comparisonsnew}
\end{figure}

\begin{figure}[h!]
    \centering
        \includegraphics[width=\linewidth]{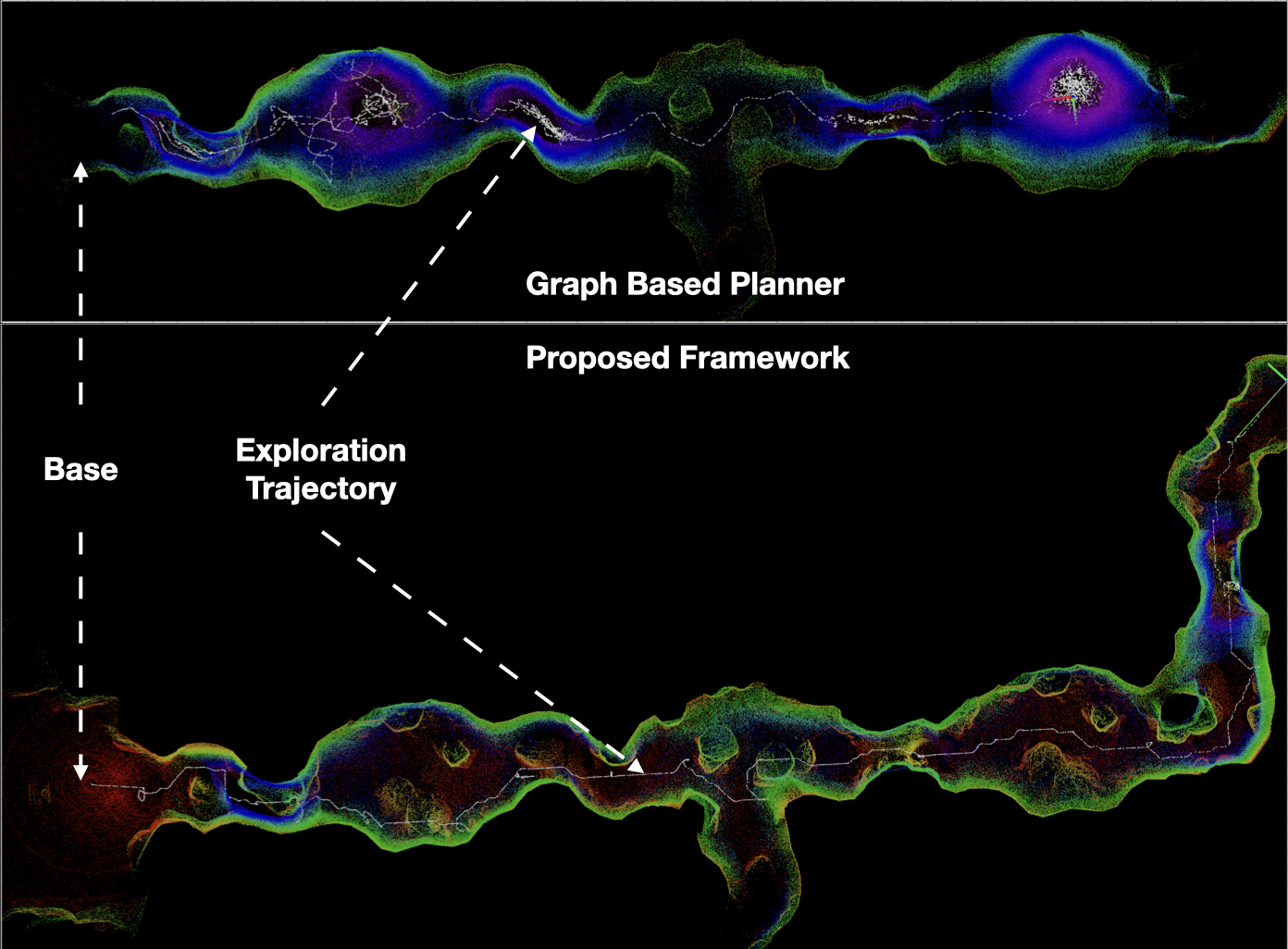}
    {\caption{Explored lava tube and tracked trajectory of MCQ using the proposed autonomy architecture. Compared against the graph based planner. The sampling based planning approach shows instances of loops in one place and in contrast, the proposed framework allows MCQ to efficiently explore new areas.}}
    \label{fig:fullruncompare}
\end{figure}

{The proposed approach is also tested intensively in simulated exploration missions and is compared against state of the art global exploration planner for validating the energy efficient nature of the proposed scheme. The exploration volume gain and ground covered by the MCQ is presented in \autoref{fig:comparisonsnew}. The same is visible in~\autoref{fig:fullruncompare} where using the proposed approach the MCQ explores larger part of the lava tube.}
The video of the proposed exploration mission and multimedia material can be found at \url{https://youtu.be/9mc0UzAJC3U}.

\section{Related Works}\label{sec:relatedworks}

Mars exploration have been prime interest in terrestrial exploration of solar system bodies due to its place in solar system and relatively easy access to send autonomous robots for exploration and surveying of Martian terrain. As mentioned earlier in \autoref{sec:introduction}, the Mars helicopter's ground breaking success on first powered flight in thin Martian atmosphere has attracted ideas and designs for future Unmanned Aerial Vehicles (UAVs) based surveying of Martian terrain. {In~\citep*{tzanetos2022ingenuity} A detailed study on the Mars helicopter's mission profiles, operations, anomalies of the design as well as lessons learned are presented based on retrieved flight data. The studies presented in\citep*{tzanetos2022ingenuity} shows that the helicopter as an aerial scout has helped the Perseverance rover in multiple instances to refine its planned route to go around rocks and boulders that could be a potential hazard to mission. The aerial images taken by the helicopter are laid over low resolution images from satellite in Mars orbit to further evaluate future path planning and steps for the rover. More interpretations of the flight data and details on Ingenuity's operations are presented in \citep*{grip2022flying}, \citep*{alibay2022operational} and \citep*{ehrenfried2022ingenuity}. More recent designs for future aerial exploration of Martian terrain are presented in \citep*{pipenberg2022conceptual} and \citep*{lorenz2022planetary} that include designs inspired from Ingenuity but with the idea of hybrid motion of aerial exploration as well as wheeled locomotion for sample retrieving purposes. Although the Mars helicopter has surpassed exprectations in terms of performance and operations for a technology demonstrator vehicle, the lessons learned from Ingenuity also show that it lacks advanced autonomy architecture to perform exploration and surveying independent of the commands from mother ship rover. In~\citep*{tzanetos2022future} a design for a Mars Science Helicopter vehicle is proposed with the focus on advancing the operations carried out by Ingenuity but with a bigger science payload and a longer range of flight. The vehicle proposed in~\citep*{tzanetos2022future} is a hex-coaxial based multi rotor vehicle designed to operate in Martian atmosphere. However, from the best of our knowledge, the autonomy enabling sensor suite and control architecture proposed in~\citep*{tzanetos2022future} is still using the same duo of Laser Range Finder (LRF) and down facing gray-scale navigation camera as Ingenuity \citep*{bayard2019vision} for guidance and navigation purpose. In contrast, in our previous work~\citep*{patel2021design}, ~\citep*{patel2022design} a Mars coaxial quadrotor vehicle named MCQ is proposed with advanced sensor suite for navigation in subsurface void or potential lava tube on Mars.} The subterranean areas are not always well lit and therefore, vision-only based navigation might be challenging to back a sophisticated on board GNC architecture. The autonomy architecture on Ingenuity helicopter also relies on PID based casced control for low level actuation and monocular vision + inertial measurements unit based localization for performing autonomous flights. {The design aspect of rotor crafts for Mars condition is comprehensively discussed in the literature~\citep*{johnson2020mars}, however it lacks a dedicated advanced form of control architecture for achieving higher levels of autonomy. In order to address the shortcomings of the state of the art vehicle, a specially designed Coaxial Quadrotor for Martian conditions and having a nonlinear model predictive controller, is utilized for simulating a fully autonomous exploration mission as it is proposed in this article. Moreover, a PID based control might yield acceptable performance in position control in open areas however, the targeted application for future aerial scouts are subterranean lava tubes. Thus, advanced trajectory tracking, reactive navigation and control architecture is absolute necessity for an autonomous vehicle to be deployed in completely unknown areas. In response we propose a complete exploration-planning-control architecture that considers energy efficiency along with safety of the vehicle as ultimate constraints to simulate a realistic lava tube exploration mission in Martian conditions.}

In the context of subterranean exploration on another planet, there are multiple driving factors to deploy autonomous robots to complete the task. The factors include achieving a completely autonomous exploration and returning with accurate 3D reconstructed map of the explored area while still efficiently utilizing the available flight time by rapidly exploring unknowns. {Since robotic exploration of unknown area is a research direction that has been pursued with many different approaches, a frontiers based approach has been fundamental. The main idea of autonomous robotic exploration is to solve the Next Best View (NBV) problem subject to adequate constraints that focus on certain flight behaviour. For example, in this work the motive of the exploration and planning architecture is to maximize exploration volume while efficiently utilizing the energy resource. The exploration approaches in literature can be classified in two high level strategies named as frontiers based and sampling based approaches.} In robotics, a frontier is defined as the boundary between unknown and free space around the robot as it was initially presented in ~\citep*{yamauchi1997frontier} for single robot and in~\citep*{yamauchi1998frontier} for multiple robots. In the initial work of frontiers based exploration, a closest frontier from the mobile robot was chosen to navigate to followed by a 360$^{\circ}$ sensor sweep to detect further frontiers. Most of the modern frontiers based approaches rely on similar approach (closest frontier) to select next goal for the robot to visit with some additional modifications for example frontiers clustering. The frontiers based approaches are further studies in ~\citep*{julia2012comparison} and \citep*{holz2010evaluating} for comparisons against various other exploration approaches. {A frontiers clustering based 3D exploration tool was proposed in \citep*{zhu20153d} that uses a quadrotor equipped with a stereo visual camera for exploration. The approach proposed in~\citep*{zhu20153d} is based on selecting a closest frontiers cluster from robots position for incremental exploration. However selecting a closest frontier based approach in not optimal for the application discussed in this work due to the fact that the closest frontier cluster might not always lie in front of the aerial vehicle and therefore requiring additional energy usage for subsequent yaw movements to navigate towards closest frontier cluster. In subterranean exploration local way point selection planner is presented in~\citep*{patel2022fast} where authors propose a local planner based on computing safe look ahead poses for aerial robot while reactively navigating in obstructed areas. More works in this direction also include \citep*{fraundorfer2012vision} which is based on vision-only based exploration-mapping scheme that also utilize Micro Aerial Vehicle (MAV) to explore using continuously updating frontiers. A generic exploration approach for UAVs is also discussed in \citep*{cieslewski2017rapid} that uses rapid frontiers selection approach in which frontiers corresponding to high flight velocities are selected ahead of the UAV. Such rapid frontier selection approach has shown relatively superior exploration gain compared to the receding horizon next best view planner \citep*{bircher2016receding}. However, when there exist no frontier ahead of the UAV, the method in~\citep*{cieslewski2017rapid} switches to classical frontiers based approach which again disregard the constraint of energy efficiency. Further studies in \citep*{zhou2021fuel} shows that rapid flight driven frontier selection approach often gets stuck in maze or multi branches structure scenarios due to switching to classical frontiers based approach at multiple instances.} More theoretical ideas of exploration including  Stochastic Differential Equation (SDE) based exploration approach is presented in \citep*{shen2012autonomous} where expansion of system of particles with Newtonian dynamics for evolution of SDE is simulated. The regions with high particle expansion are portrayed to be unknown areas for the MAV to explore. 

{The second mainly researched direction in exploration is sampling based approaches that often integrate exploration and path planning as coupled problem to be solved. Such methods are often based on computing local goals for exploration while maximizing information gain and at the same time planning paths in order to minimize distance travelled, actuation cost or similar metrics\citep*{lindqvist2021exploration}. These sampling based approaches based on Rapidly-exploring Random Trees (RRT) structures are extensively studied in both simulations and real life experiments as presented in \citep*{bircher2016receding}, \citep*{pito1999solution} and \citep*{dang2020autonomous}. In~\citep*{sun2022ada} a hybrid approach is discussed where, an adaptive frontier detector for rapid exploration based on local sampling in RRT is presented. The approach has bench marked the exploration in indoor, forest as well as multi story building for improved exploration gain and sampling rate which is a challenge in RRT based methods. However, multiple instances have been seen where the UAV goes back and forth in local space while exploring and this is common challenge in any sampling based approach. In~\citep*{dharmadhikari2020motion} a motion primitives based exploration-planning approach is presented where next exploration goal and subsequent path is planned such that trajectory tracking can be improved for local exploration. The work of \citep*{dharmadhikari2020motion} is excellent candidate to be inline with the focus of energy efficient exploration however, the method often gets stuck in local minima when exploration gain is low and also due to the lack of dedicated global re-positioning scheme that could be triggered automatically when local exploration is insufficient as it will be presented in this work. Apart from local exploration-planning methods, a global graph based planner is also presented in~\citep*{dang2019graph}, \citep*{dang2020graph}. So far the graph based planner is the state of the art method in subterranean exploration of unknown environments which has dedicated global re-positioning strategy. Therefore in this work we have presented the comparisons of our work with \citep*{dang2020graph} to evaluate exploration metric.} Apart from sampling and frontiers based approaches, there exist also hybrid approaches. An information driven frontier exploration method for MAVs, which uses a hybrid approach between control sampling and frontier based is presented in~\citep*{dai2020fast}. {The graph based strategy is also extended in terms of multi robot teamed exploration of subterranean caves and tunnels as part of DARPA subterranean challenge is presented in \citep*{kulkarni2022autonomous}.} Exploration of unknown environments are also extended to legged or ground robots. Probabilistic Local and Global Reasoning on Information roadMaps (PLGRIM) as presented in \citep*{kim2021plgrim}, discusses a hierarchical value learning strategy for autonomous exploration of large subterranean environments. The methodology presented in~\citep*{kim2021plgrim} uses a hierarchical learning to address local and global exploration of large scale environments while focusing on near optimal coverage plans. A Frontloaded Information Gain Orienteering Problem (FIG-OP) based strategy is presented in~\citep*{peltzer2022fig} that uses topological maps to plan exploration paths in fixed time budget exploration scenario. The method presented in~\citep*{peltzer2022fig} is tested with ground robots in multi-kilometers subterranean environment targeted at time constrained exploration missions.

{Apart from the exploration and planning modules, the third necessary component of the autonomy architecture is a control scheme for the vehicle. The most basic form of position control utilized by autonomous aerial vehicles is PID control. The Mars helicopter also has nested low level control that is based on a simple PID architecture. However, the position control using a PID method does not always yield accurate tracking due to the overshoots in non smooth trajectories. Other commonly used form of control strategies include Linear Quadratic Regulator (LQR) and Model Predictive Control (MPC) based position and trajectory tracking controls. Initial works on model predictive control for quadrotor class of vehicle include \citep*{bangura2014real}, \citep*{abdolhosseini2013efficient}, \citep*{alexis2014trajectory}, \citep*{chikasha2017adaptive}. More MPC based navigation scheme in subterranean environments are proposed in \citep*{mansouri2020subterranean} and \citep*{lindqvist2020non}. Separated from quad rotor vehicle control, recent studies have presented more advanced form of control for space vehicles as presented by authors in \citep*{alsaade2022neural} for fixed-time attitude tracking control that utilizes berrier Lyapunov functions with neural network. The extended form of neural adaptive fault-tolerant control strategy is presented in \citep*{alsaade2022indirect}. The authors validate the proposed indirect neural approximation based position and attitude tracking in presence of unknown disturbances through extensive simulation studies to track freely tumbling target in an orbit. Spacecraft control under uncertainties is a challenging research direction pursued by novel control applications such as attitude stabilization and vibration suppression for a flexible spacecraft as described by authors in \citep*{yao2022neural}. A neural integral sliding mode based control scheme is proposed by authors in \citep*{yao2022indirect} that establishes a finite time attitude tracking of a spacecraft in presence of unknown uncertainties and disturbances. The control method presented in \citep*{yao2022indirect} also shows reduced computational complexity of the control scheme due reduced number of learning parameters to be tuned online. In \citep*{alsaade2022new} a neural network based optimal mixed control scheme (targeting  unknown uncertainties and disturbances) for a custom modified UAV is proposed.}

{The related works in the field of autonomously exploring aerial vehicles have laid the foundation for future aerial exploration of Mars. However, the methods somehow lack the idea of preserving energy or efficient energy utilization while exploring. Thus, through this work we present novel autonomy framework to contribute towards solving challenge of energy efficient autonomous exploration of potential lava tube on Mars. Based on the related works, the key contributions of this work are presented in \autoref{sec:contributions}.}

\section{Motivation}\label{sec:motivation}

{Martian terrain exploration with aerial robots is becoming emerging research direction because it addressed the limitations of rovers in terms of traversability and navigation speed, demanding novel solutions by subdisciplines of Artificial Intelligence including navigation, exploration, mapping, perception, planning and control. The aerial surveying further complements the idea of exploration by not only being the aerial scouts for rovers but also have the ability to enter potential subsurface voids or lava tubes on Mars which are difficult or untraversable terrains for rovers. This work is inspired from the successful flight attempts of Ingenuity, the Mars helicopter which was sent to Mars along with the Perseverance rover. The valuable research data collected from multiple flights of Ingenuity on Mars have proved that powered flight in thin Martian atmosphere is possible. However, the lessons learned from Ingenuity also show that flight time of such vehicle is extremely limited and making the best out of limited flight time directly relates with how efficiently the vehicle can utilize the available energy from power source.}

{It is a major challenge for multi rotor vehicles to produce lift in thin Martian atmospheric conditions. In design of such vehicles, this challenge is addressed with combination of two factors that directly affect the lift capacity and flight time of vehicle. The optimal trade-off between how fast the rotors should spin and how big the rotors should be to generate enough lift derives the flight time for such vehicles. The first factor (rotor speeds) directly relates to the energy utilization from batteries and thus proportionally the exploration volume and distance covered by a vehicle in the limited flight time. The state of the art methods discussed in \autoref{sec:relatedworks} are mainly classical frontiers or sampling based. The classical frontiers based methods utilize a frontier selection based on minimum distance function which result in multiple yaw movements for the aerial vehicle because the closest or minimum distance frontier might not always lie in front of the vehicle and therefore, exploration based on classical frontiers based method result in multiple instances where aerial vehicle utilizes more energy for yaw correction and resulting in low exploration volume gain in fixed time budged based missions. Moreover in the classical approaches, the vehicle back tracks previously visited positions in cases where the vehicle reaches dead end in order to globally reposition to a partially seen area to continue exploration in different branch of cave or lava tube. This often result in longer paths to go back to junction and therefore proportionally utilizing more energy to achieve same task. In contrast the proposed scheme utilizes a risk aware expendable grid based planning method that computes shortest-collision-free paths to navigate towards partially seen area in case of dead end. On the other hand, sampling based approaches such as RRT inspired methods often lack the ability of rapidly computing the future exploration goals and therefore, often require the vehicle to hover at one place for a fraction of time while the algorithm computes next goal and path. This result in \textit{explore-stop-plan-explore} flight behaviour which again disregard the idea of efficiently utilizing the available flight time to maximize exploration.}

{The future Martian rotor crafts such as Mars Coaxial Quadrotor (MCQ) requires intelligent rapid exploration strategies that prevents the vehicle to hover at one place while planning future frontiers and path to the frontiers as well as preventing motor saturation by minimizing yaw torque along vertical axis. Therefore, the main motivation of this research work is to present a novel exploration-planning autonomy architecture for MCQ that addresses the efficient energy utilization for MCQ vehicle in a simulated Martian lava tube exploration scenario. As it will be presented in next sections, we propose energy preserving frontiers based exploration approach along with risk aware planning, reactive navigation and control scheme for aerial exploration of a potential lava tube on Mars.}

\section{Contributions}\label{sec:contributions}

The exploration-planning architecture of this work is part of the development efforts within the COSTAR team~\citep*{agha2021nebula}, related with the DARPA Sub-T competition~\citep*{subtworld}, while the applicability of the proposed scheme is extended for planetary caves and lava tubes as subterranean environments. The simulated exploration mission presented in this article is motivated from our previous work on design and model predictive control of a coaxial quadrotor to operate in Mars conditions. The MCQ used for simulation is presented in~\citep*{patel2021design}. Based on the related works and existing approaches on aerial robots for planetary exploration, the novel contributions of this work are listed below.

{The first contribution of this work is in relation with addressing the main challenge on efficiently utilizing the energy resource of Mars Coaxial Quadrotor (MCQ) while exploring. In response, this work establishes an exploration and planning scheme for the MCQ that takes into consideration minimizing the yaw movement of MCQ as well as computing next frontier and planning the path to next frontier on the go. The back bone of the proposed exploration-planning architecture is derived from the \textit{Look-Ahead Move-Forward} nature of flight. In order to rapidly move forward in an unknown environment, the proposed work also establishes a two layer safe frontier detection and selection approach. The first layer of exploration computes the occupancy information of the surrounding, stored in an octree data structure format and detects safe frontier points for the MCQ such that the MCQ can navigate to such frontier in nearly straight path in a collision free manner. The second layer deals with selecting an optimal frontier subject to constraints related with preventing unnecessary iterations for fast processing and forcing the MCQ to move forward as long as there is unknown space ahead of it by bifurcating the frontiers into directly and indirectly accessible frontiers. As long as there isn't a dead end ahead of the MCQ, it prioritizes a frontier from directly accessible frontiers set that require minimum yaw movement corrections. In multi rotor systems, yaw movement tend to saturate the motors by dawning excess energy from batteries and and therefore, exploring with minimal yaw movements considerably helps the MCQ in efficient energy utilization. The state of the art sampling based approaches for planning, work in \textit{Explore-Stop-Plan-Explore} manner. Moreover the multi rotor vehicle's energy is best utilized while flying at nearly constant speeds while avoiding hovering in one place. In contrast, the in this work we propose an exploration-planning architecture that rapidly computes frontiers and plans paths while exploring forward. In simple words, avoiding the hovering in one place to plan next steps.}

{There are moments in exploration of unknown subterranean environments, when MCQ might reach a dead end of a lava tube. In such scenario the MCQ iterates through the set of Indirectly accessible frontiers and computes for an optimal frontier subject to minimizing energy usage to navigate to such frontier. We propose a cost based indirectly accessible frontier that takes into consideration yaw correction cost, euclidean distance as well as height difference to avoid saturating the motors and to help in efficiently utilizing energy resource.} 

{Apart from the exploration algorithm, the proposed work also utilizes a risk aware expendable grid based planning $D^{*}_{+}$ method along with Artificial Potential Fields (APF) for redundant collision avoidance as well as a Nonlinear Model Predictive control path following. We integrate the exploration, planning as well as reactive navigation and control modules into one autonomy framework designed specifically to address challenges of autonomously exploring a lava tube structure beneath Martian surface. The novelty of the proposed autonomy architecture stems from the redundancy and dependency less overall work flow within the systems. The framework is evaluated through intensive simulation studies and comparisons with state of the art sampling based approach. The simulations are carried at true Mars atmospheric conditions such as Martian air density, surface pressure and gravity to evaluate the method in a truly realistic martian lava tube environment. We have also constructed a custom Martian lava tube environment which represent potential martian underground terrain for simulation purpose. We have made the environment publicly available for fellow researchers to use.}

The rest of the article is structured as follows. \autoref{sec:coax_quad} provides details on the design and torque based dynamic model of a custom Martian Coaxial Quadrotor used for simulation purposed.  \autoref{sec:IFE} presents the proposed frontier based exploration approach with focus on energy preserving frontier generation. In \autoref{sec:sim_results}, simulation results are presented that prove the efficacy of the proposed scheme. Finally \autoref{sec:conclusions} provides a discussion with concluding remarks and future directions of the proposed work.
%
\section{Mars Coaxial Quadrotor (MCQ)}
\begin{figure}[h!]
  \centering
    \includegraphics[width=\linewidth]{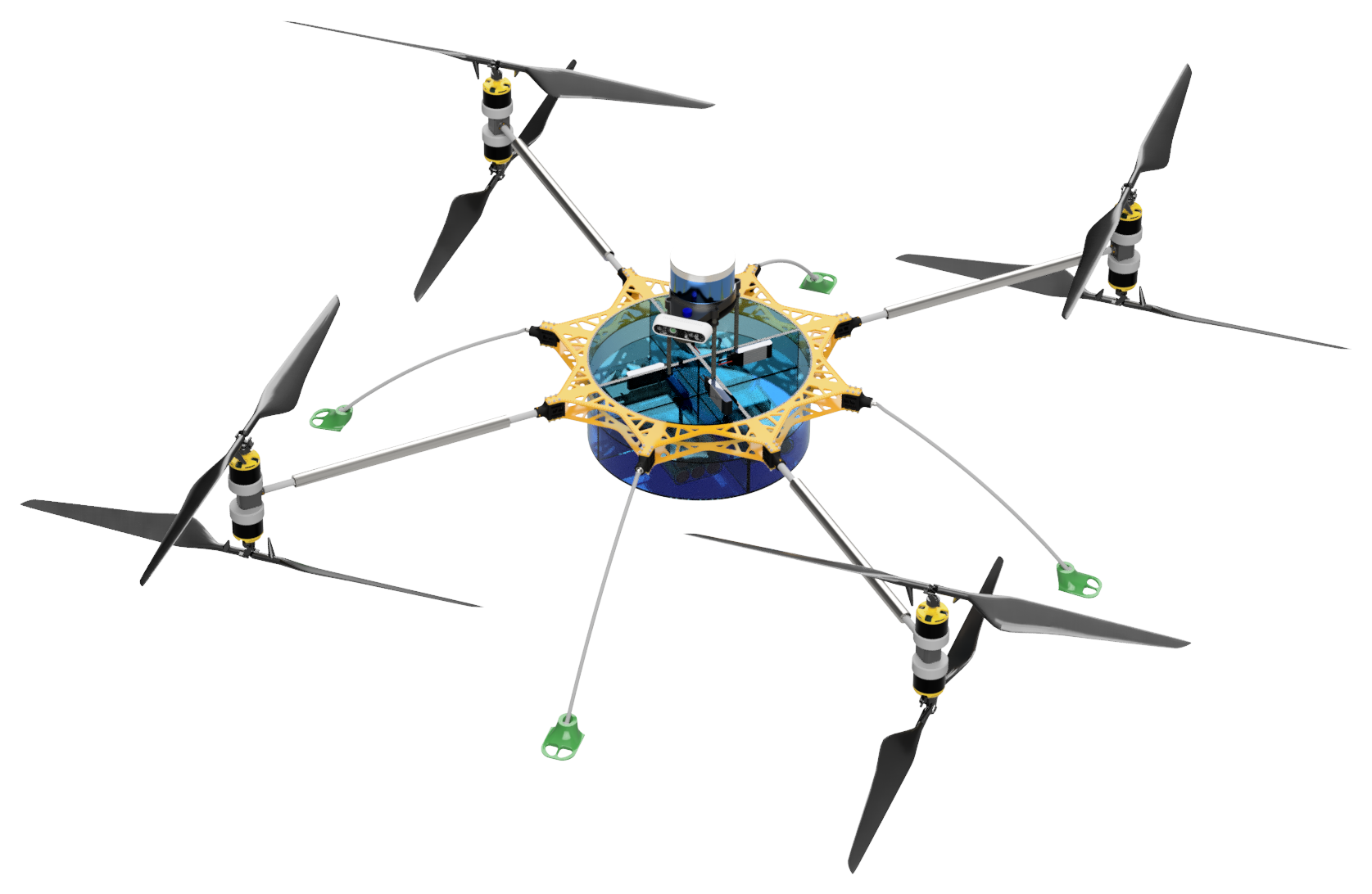}
    {\caption{Mars Coaxial Quadrotor (MCQ) used for the exploration mission.}}
  \label{fig:marcoax}
\end{figure}

{\label{sec:coax_quad}}
%
The Mars coaxial Quadrotor design optimizations are specifically done to operate in the Mars thin atmosphere. The detailed analysis and modelling of the Mars coaxial Quadrotor is presented in our previous work~\citep*{patel2021design}. Due to the reason that the Mars coaxial Quadrotor is designed to operate in thin atmosphere, it's rotors have to spin at considerably high RPM in order to produce the required thrust. Since the high RPM of the rotors by default consume considerably high power from the batteries, it is important to optimally utilize resources in terms of electrical power in an exploration scenario. The model of the Mars coaxial Quadrotor is presented in \autoref{fig:marcoax}.
The proposed exploration algorithm is based on minimizing the change in the direction vector by forcing to select the frontier in the field of view. The algorithm also implements a cost function based selection process when there is no frontier lying in the field of view, which allows the MCQ to reject a frontier for which the MCQ has to deviate significantly from the current direction. Due to this novelty, the MCQ is able to maintain a stable flight, avoiding aggressive maneuvers which eventually efficiently uses the limited energy available from batteries. Due to the fact that Mars has thin atmosphere and low surface pressure, the rotors of MCQ need to spin considerably faster and thus, it is important to optimize the energy consumption by the Mars Coaxial Quadrotor.

\begin{figure}[h!]
  \centering
    \includegraphics[width=\linewidth]{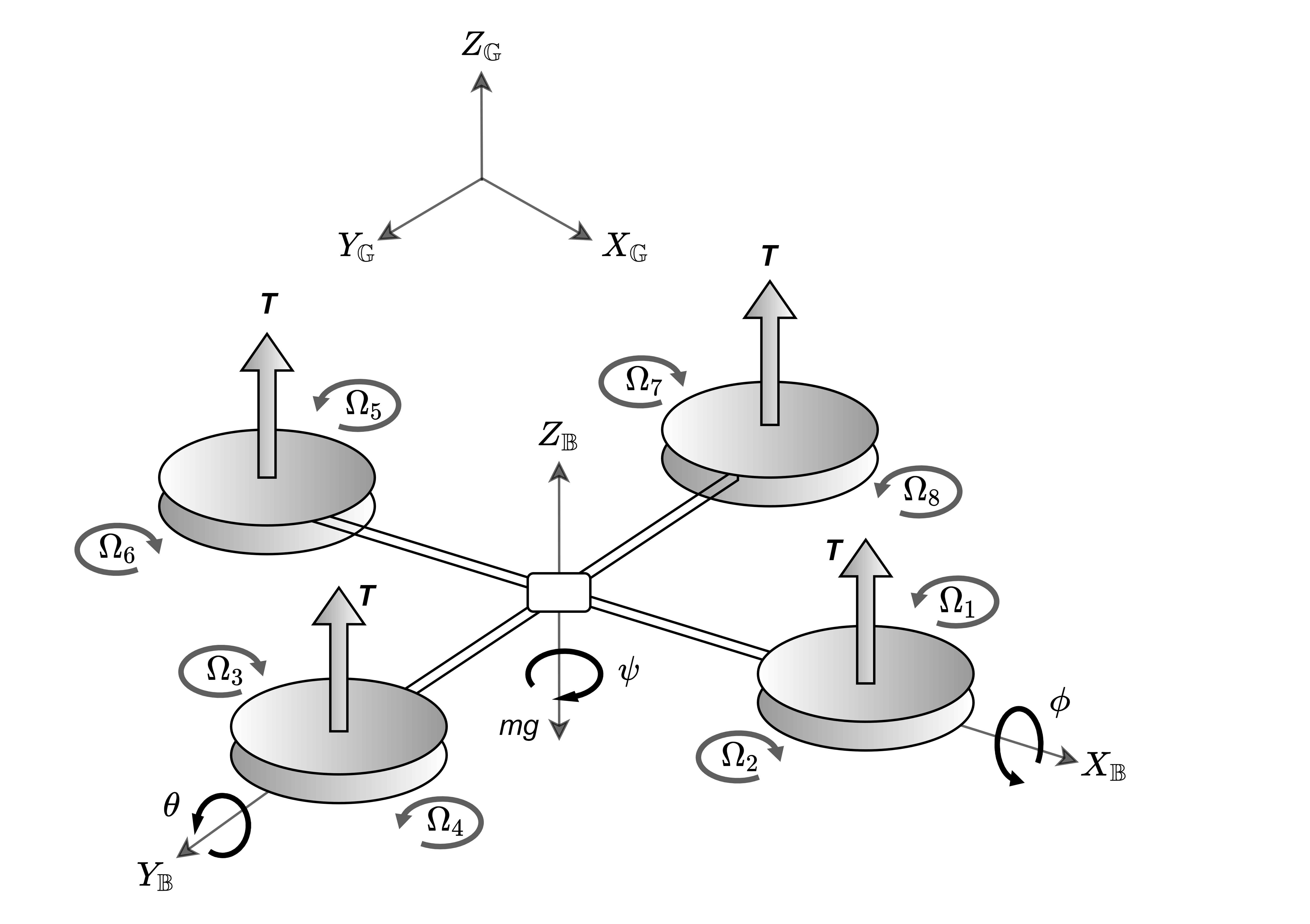}
    {\caption{Coaxial quadrotor system representation.}}
  \label{fig:mcoax}
\end{figure}

The torque based model of the coaxial Quadrotor is presented in Equation~\ref{allocation}. 
$T$ is the thrust force in the $Z$ direction, $\tau_{\phi}$, $\tau_{\theta}$ and $\tau_{\psi}$ are the roll, pitch and yaw torques respectively along the $X_{\mathbb{B}}, Y_{\mathbb{B}}, Z_{\mathbb{B}}$ axis

\begin{equation*}
    \begin{psmallmatrix}
    T \\ \tau_{\phi} \\ \tau_{\theta} \\ \tau_{\psi}
    \end{psmallmatrix} =
    \begin{psmallmatrix}
    K_{T}& K_{T}& K_{T}& K_{T}& K_{T}& K_{T}& K_{T}& K_{T} \\ 0& 0& dK_{T}& dK_{T}& 0& 0& -dK_{T}& -dK_{T} \\ dK_{T}& dK_{T}& 0& 0& -dK_{T}& -dK_{T}& 0& 0 \\ -K_{D}& K_{D}& -K_{D}& K_{D}& -K_{D}& K_{D}& -K_{D}& K_{D}
    \end{psmallmatrix}
    \begin{psmallmatrix}
    \Omega_{1}^{2} \\ \Omega_{2}^{2} \\ \Omega_{3}^{2} \\ \Omega_{4}^{2} \\ \Omega_{5}^{2} \\ \Omega_{6}^{2} \\ \Omega_{7}^{2} \\ \Omega_{8}^{2}
    \end{psmallmatrix} 
 \end{equation*}

\begin{equation}
    = A 
    \begin{psmallmatrix}
    \Omega_{1}^{2},\Omega_{2}^{2}...,\Omega_{8}^{2}
    \end{psmallmatrix}^T
    \label{allocation}
\end{equation}
In Equation~\ref{allocation}, the matrix $A \in \mathbb{R}^{4\times8}$ is the allocation matrix responsible for producing specific control inputs based on desired state of the model, while $K_T$ and $K_D$ are the thrust and drag coefficients and $\Omega_{i}^2$ is the rotor speed in $rad/s$. The approximate values for $K_T$ and $K_D$ are presented in Table \ref{table:1}.

\begin{table}[h!]
\small
\renewcommand{\arraystretch}{1.3}
\caption{\bf Mars Coaxial Quadrotor and simulation model parameters}\label{table:1}
\centering
\begin{tabular}{|c|c|} 
 \hline
\bfseries Simulation Parameter & \bfseries Value \\  
 \hline\hline
 
Rotor inertia, $J$ ($Kg \cdot m^3$) & 4.240$e^{-4}$ \\ 
Torque constant \textit{$k_t$} ($Nm/A$) & 0.010$e^{-3}$ \\
Thrust coefficient $K_T$ & 0.60 \\
Drag coefficient $K_D$ & 0.20$e^{-3}$ \\
Density, $\rho$ (Kg/$m^3$) & 0.017 \\ 
Static Pressure \textit{p} (Pa) & 720 \\
Operating Temperature \textit{T} (K) & 223 \\
Gas constant \textit{R} ($m^{2}/s^{2}$/K) & 188.90 \\
Dynamic viscosity \textit{$\mu$} (Ns/$m^2$) & 1.130$\cdot10^{-5}$ \\
Gamma \textit{$\gamma$} & 1.289 \\  
 
 \hline
\end{tabular}
\normalsize
\end{table}

\section{Incremental Frontiers Based Exploration}\label{sec:IFE}
%
The proposed extended frontier based method uses occupancy grid maps as a mapping algorithm, which can generate a 2D or 3D probabilistic map. A value of occupancy is assigned to each cell that represents a cell to be either free or occupied in the grid. Suppose $m_{x,y}$ is the occupancy of a cell at $(x,y)$ and $p(m_{x,y}\ |\ z^{t},x^{t})$ is the numerical probability, then by using a Bayes filter~\citep*{chen2003bayesian} the odds of the cell to be occupied can be denoted as:
\small
\begin{equation*}
    \frac{p(m_{x,y}\ |\ z^{t},s^{t})}{1-p(m_{x,y}\ |\ z^{t},s^{t})} =  \frac{1-p(m_{x,y})}{p(m_{x,y}}\times\\ \frac{p(m_{x,y}\ |\ z^{t-1},s^{t-1})}{1-p(m_{x,y}\ |\ z^{t-1},s^{t-1})}
\end{equation*}
\normalsize
where $z^t$ represent a single measurement taken at the location $s^t$. In order to construct a 3D occupancy grid an existing framework OctoMap~\citep*{hornung2013octomap} based octree is formulated in this work. In this case each Voxel is subdivided into eight Voxels until a minimum volume defined as the resolution of the octree is reached. In the octree if a certain volume of Voxel is measured and if it is occupied, then the node containing that Voxel can be initialized and marked occupied. Similarly, using the ray casting operation, the nodes between the occupied node and the sensor, in the line of ray, can be initialized and marked as free. This leaves the uninitialized nodes to be marked unknown until the next update in the octree. The estimated value of the probability $P(n\ |\ z_{1:t})$ of the node $n$ to be occupied for the sensor measurement $z_{1:t}$ is given by:
\small
\begin{equation*}
    P(n | z_{1:t}) = [1 + \frac{1-P(n|z_{t})}{P(n|z_{t})} \frac{1-P(n|z_{1:t-1})}{P(n|z_{1:t-1})} \frac{P(n)}{1-P(n)}]^{-1}
\end{equation*}
\normalsize
where $P_{n}$ is the prior probability of node $n$ to be occupied. 
Let us denote the occupancy probability as $P^{o}$. 
$$
p^{Voxelstate}=\begin{cases}
                    Free, & \text{if $P^{o} < P_{n}$}\\
                    Occupied, & \text{if $P^{o} > P_{n}$} \\
                    Unknown, & \text{if $P^{o} = P_{n}$} \\
                \end{cases}
$$                

The proposed exploration algorithm is made up of three essential modules, namely the frontier pose generation, accessible frontier selection and global re positioning. The first module converts a point cloud into a Voxel grid $V$ defined as $V=\{\Vec{x}\}$ based in the Octomap framework discussed earlier. The next module generates frontiers based on the occupancy probability function for each Voxel corresponding to each sensor measurement. The frontier points are fed into the next module, which evaluates each frontier based on a cost function and selects an optimal frontier point. This point is published to the Mars Coaxial Quadrotor (MCQ) controller as a temporary goal point to visit and the overall process continues for each sensor scan until no frontier is remaining. 

The classical definition of frontiers considers a cell in the Voxel grid to be a frontier if at least one of it's neighbour cell is marked as unknown. In the real scenario, where the area that needs to be explored is vast (for example, a lava tube stretching for multiple kilometers on Mars), it is practically inefficient. Therefore in the proposed algorithm the number of unknown or free neighbours can be set by user before the start of the exploration that reduces the overall computational complexity by a significant margin. Another modification compared to classical frontier approach is made by not allowing any neighbour to be an occupied Voxel. This also allows constrained frontier generation such that confined narrow passages (smaller than the safety margin for the MCQ to traverse in) can be avoided and instead other areas can be prioritised. 

The third layer of frontier refinement is set as if a frontier is detected within the sphere of radius $r$ (set by user at the start of exploration), then the node is marked as free. The sphere of radius $r$ is assumed to be known space around the drone such that $r < R$ where  $R$ is the sensor range configured in the octree formulation. This helps in limiting the undesired change in vertical velocity of the MCQ if the frontier is detected above or below the MCQ. The proposed algorithm uses a 3D Lidar to get the point cloud and because of the limited vertical field of view of the 3D Lidar, the above mentioned layer of refinement helps in an overall stable operation. 
\begin{algorithm}
\small
\caption{Frontier Point Detection}
 \hspace*{\algorithmicindent} \textbf{Input : } $Current octree,$ \\
    \hspace*{1.8cm}$n$  : Number of unknown or free neighbours required \\
    \hspace*{1.8cm}$r$ : known boundary around MCQ \\
 \hspace*{\algorithmicindent} \textbf{Output : } $\mathcal{F}$
\begin{algorithmic}[1]
\FORALL{$Cell : UpdatedCells()$}
\IF{$Cell.is.Free()$}
\IF{$Cell.distance() < r$}
\STATE $i \gets 0$
\FORALL{$Neighbours : Cell.Neighbour()$}
\IF{$Neighbour.is.Occupied()$}
\STATE i = 0;
\STATE $break;$
\ELSE
\STATE $i \gets i + 1$
\ENDIF
\ENDFOR
\IF{$i >= n$}
\STATE $\mathcal{F}.add(Cell);$
\ENDIF
\ENDIF
\ENDIF
\ENDFOR
\end{algorithmic}
\normalsize
\label{generation}
\end{algorithm}

\begin{figure}[h!]
  \centering
    \includegraphics[width=\linewidth]{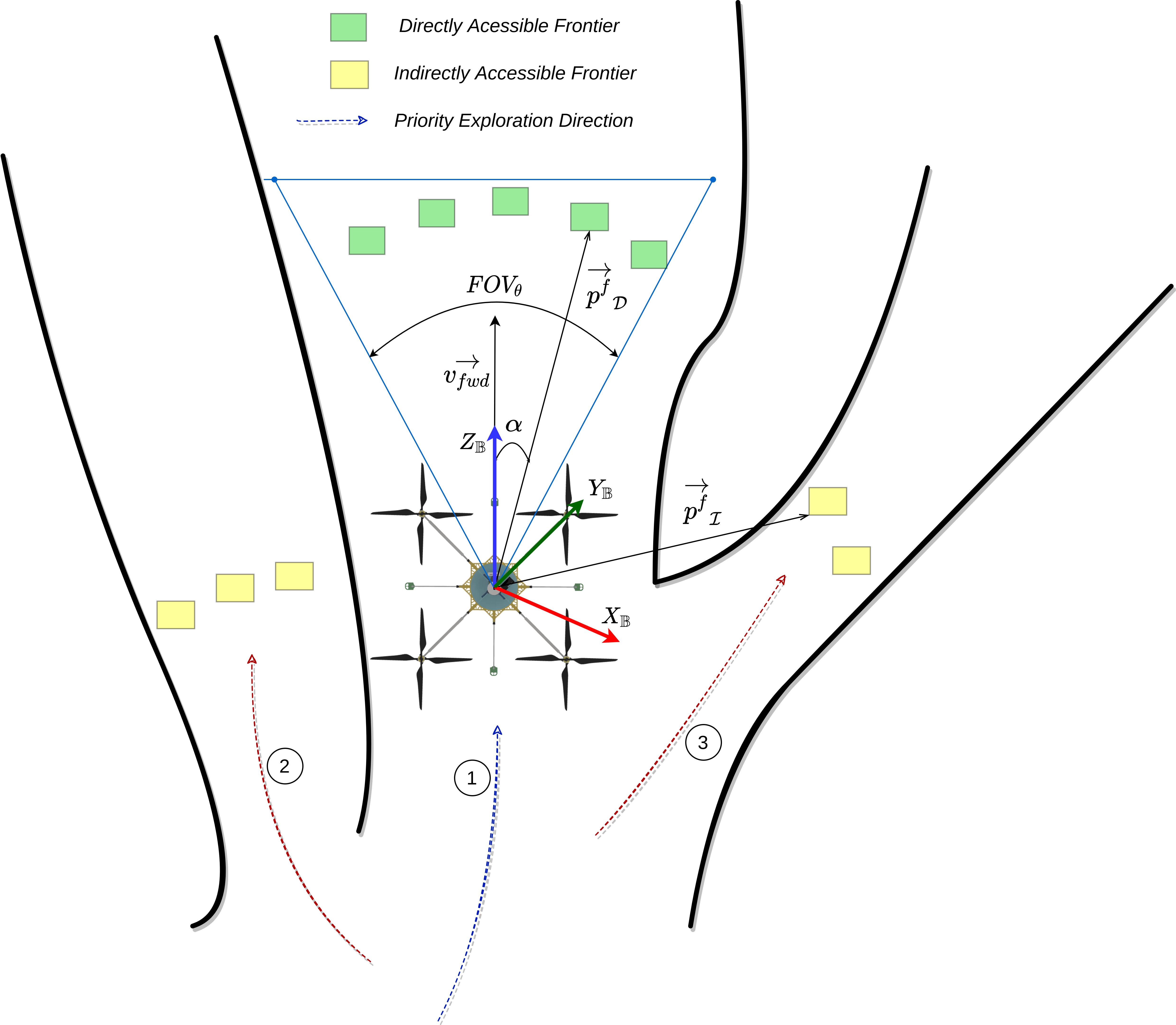}
    {\caption{Directly and Indirectly accessible frontiers classification. The motive behind bifurcating frontiers in two sets is to prioritize unknown space ahead of MCQ in a potential multi branched lava tube system.}}
  \label{fig:localandglobal}
\end{figure}
The pseudo code for the frontier generation from octomap is depicted in the Algorithm~\ref{generation}. $\mathcal{F}$ represent the set of frontiers generated in Algorithm~\ref{generation}. In order to reduce the number of cells to be examined in each update, only the cells whose state (free or occupied) is changed are examined. For the selection of frontier, a directly accessible and indirectly accessible set of frontiers are defined. For a large environment it is computationally inefficient to go through each frontier, while evaluating the optimal frontier to visit. Therefore, in the proposed method, the frontiers evaluation is done in two stages. First the frontiers in the field of view in the direction of the MCQ are added to the set which contain all frontiers that are directly accessible from the MCQ's current position and we define this set as $\mathcal{D}$. The frontiers that are not in the field of view and the once that are not directly accessible are then added top the set indirectly accessible frontiers set and is denoted as $\mathcal{I}$. In the frontier selection process as presented in algorithm~\ref{selection}, the frontiers in $\mathcal{D}$ are prioritized. When there is no frontiers in $\mathcal{D}$ frontier to visit, a frontier from the $\mathcal{I}$ is selected based on the cost function. The set of valid frontier which satisfy the requirements of having $n$ unknown or free neighbours and which are out of sphere of radius $r$ as formulated previously are then added to the set of valid frontiers denoted as $\mathcal{V}$.
\begin{algorithm}
\small
\caption{Frontier Classification}
 \hspace*{\algorithmicindent} \textbf{Input : } $\mathcal{F}$, $n$, $r$ \\
    \hspace*{1.8cm}$\alpha$ : frontier vector angle  \\
 \hspace*{\algorithmicindent} \textbf{Output : } $\mathcal{D}$, $\mathcal{I}$ 
\begin{algorithmic}[1]
\FORALL{$Frontier : \mathcal{F}$}
\IF{$(Frontier.distance < r)\ or\ (Frontier.height > h_{r})$}
\STATE $i \gets 0;$
\FORALL{$Neighbours : Frontier.Neighbour()$}
\IF{$Neighbour.is.Occupied()$}
\STATE $i \gets 0;$
\STATE $break;$
\ELSE
\STATE $i \gets i + 1;$
\ENDIF
\ENDFOR
\IF{$i < n$}
\STATE $\mathcal{F}.remove(Frontier);$
\ENDIF
\FORALL{$Frontier : \mathcal{V}$}
\IF{$\alpha$ $\leq$ $\theta/2$}
\STATE $\mathcal{D}.add(Frontier)$
\ELSE 
\STATE $\mathcal{I}.add(Frontier)$
\ENDIF
\ENDFOR
\ENDIF
\ENDFOR
\end{algorithmic}
\normalsize
\label{selection}
\end{algorithm}
The pseudo code for the frontier classification, for both directly and indirectly accessible frontiers, is presented in Algorithm~\ref{selection}. $\theta$ is 2D horizontal field of view considered to classify directly accessible frontiers. In vertical direction each frontier height is checked as described in algorithm~\ref{selection} and if frontier height exceeds then maximum height difference bound then such frontier is not considered directly accessible frontier. In algorithm~\ref{selection}, $h_{r}$ is the maximum relative height at which frontiers are consideref to be accessible. Such classification also limits how much the MCQ would move in vertical direction while exploring forward. A frontier from the $\mathcal{D}$ set is selected based on a minimum angle $\alpha \in [-\pi,\pi]$ between then frontier vector $\Vec{p^{f}}$ and the MCQ's direction of travel $\Vec{V_{fwd}}$. In \autoref{fig:localandglobal}, $(X_{\mathbb{B}},Y_{\mathbb{B}},Z_{\mathbb{B}})$ represent the body fixed coordinate frame. The frontiers from Octomap are generated in the world frame but the frontier vector $\Vec{x_{f}}$ is calculated relative to the position of the MCQ. Let us denote $P^{mcq}_{k}$ as the position of the MCQ at time instance $k$, where $P^{mcq} = \{x,y,z,\psi\}$. Therefore we can compute the vector defining the direction of motion of the MCQ as,

\begin{equation}
    V_{fwd} = (P^{MCQ}_{k+1} - P^{MCQ}_{k}).normalize()
\end{equation}
As shown in \autoref{fig:localandglobal}, the angle $\alpha$ with respect to $\mathbb{B}$ can be defined as:
\begin{equation*}
    \alpha_{i} = cos^{-1}({\frac{\Vec{p^{f}_{i}}\cdotp\Vec{V_{fwd}}}{|\Vec{p^{f}_{i}}|\cdotp|{\Vec{V_{fwd}}|}}})
\end{equation*}

Thus, by selecting the frontier with a lowest value of the angle $\alpha$, the exploration can be continued with a relatively higher speed as the MCQ does not have to change the direction of motion as long as a selected frontier regulates the heading of the MCQ as it explores. When there is no frontier in the $\mathcal{D}$ set then a frontier from $\mathcal{I}$ set is selected. The selection of a frontier from $\mathcal{I}$ set can be tuned based on a cost function as shown in Equation~\ref{costfunction}
\begin{equation}
    \mathbb{C} = 
\left\vert \sum_{i=1}^n (\mathcal{W}_{\alpha}\ \alpha_{i},\ \mathcal{W}_{h}\ \Delta H_{i},\ \mathcal{W}_{d}\ d_{i}) \right\vert 
    \label{costfunction}
\end{equation}
where, $d$ is the distance between the MCQ's current position and the frontier point, $\Delta H$ is the height difference between MCQ and the frontier point, and $\mathcal{W}_{\alpha}$, $\mathcal{W}_{h}$ and $\mathcal{W}_{d}$ are weights on the angle, height and distance difference respectively. The cost for each valid frontier in the global set is calculated and the frontier with a minimum cost value $\mathbb{C}$ from Equation~\ref{costfunction} is selected as the next frontier in order to globally re position the MCQ in case of a dead end or inaccessible frontier. As discussed previously, the weights can be set before the start of the exploration, which allows where the frontier will be selected relative to the MCQ's current position in the 3D space. For example, if a high value of $\mathcal{W}_{h}$ is set, then the cost concerning a height difference is increased, resulting in a frontier higher or lower than the MCQ position in the $Z$ direction that would be less likely to be selected.

\subsection{Autonomy framework for Exploration}
\begin{figure}[h!]
  \centering
    \includegraphics[width=\linewidth]{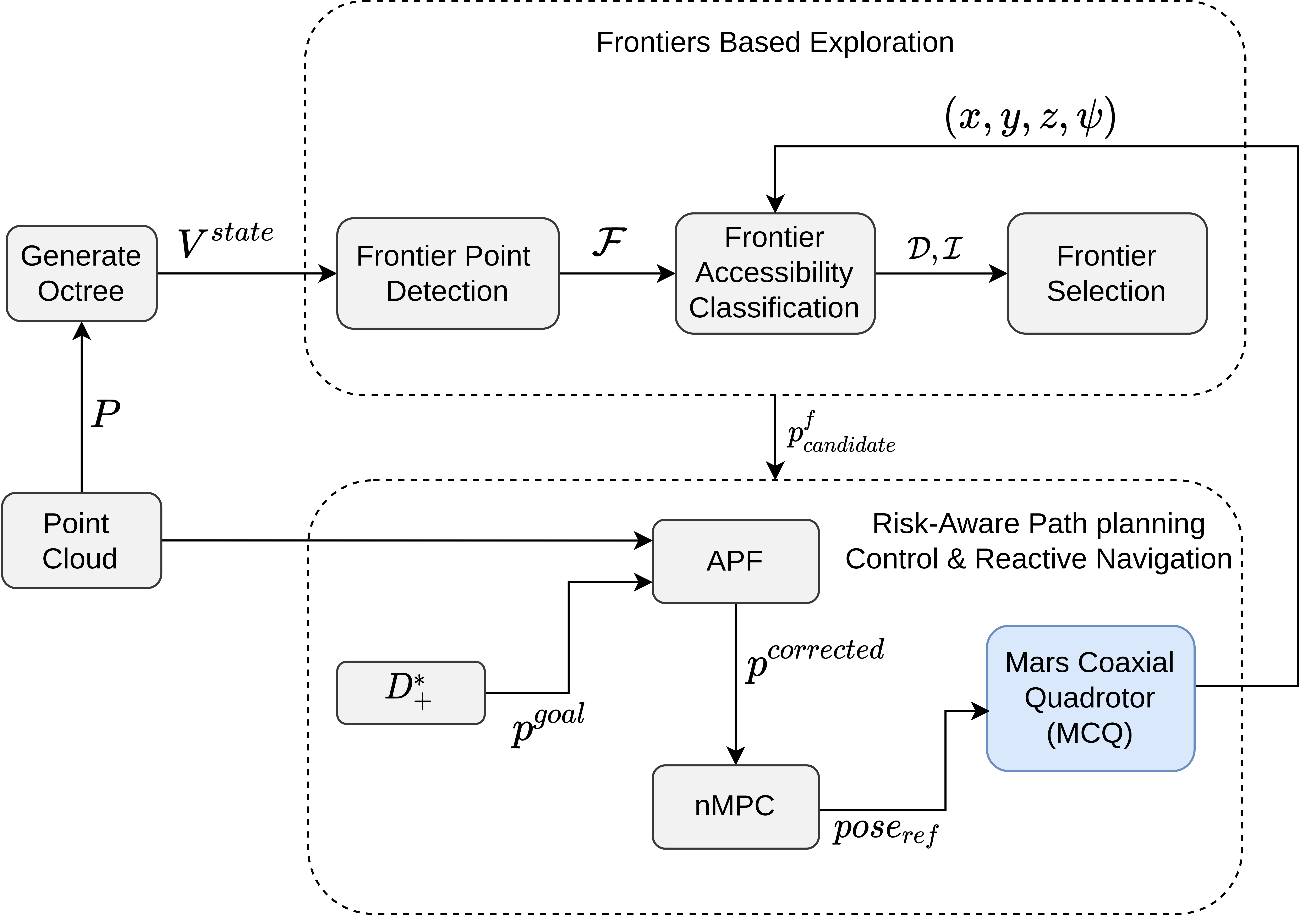}
    
    {\caption{High level autonomy framework composed of the proposed frontiers based approach along with risk-aware planning and reactive navigation-control sub modules.}}
  \label{fig:autonomy}
\end{figure}

The autonomy framework is composed of exploration, planning and control modules and is presented in \autoref{fig:autonomy}. Furthermore, the considered exploration mission relies on a 3D lidar data for frontier generation, as well as for collision avoidance. As presented in the previous \autoref{sec:coax_quad}, the nonlinear model predictive control of the MCQ is achieved by allocating rotor speeds converted from the desired thrust, roll, pitch and yaw. In \autoref{fig:autonomy} the complete framework of exploration and risk aware path planning is shown for an autonomous exploration mission in a mars like environment. The overall high-level autonomy architecture of the MCQ is divided into two main modules namely exploration and planning-control. As presented in \autoref{sec:IFE}, the exploration module takes point cloud and generates an octree based on occupancy information. The octree provides the voxel states denoted as $V^{state}$ to the frontier detection module. The process is followed by the frontier classification module which classifies the frontiers in $\mathcal{D}$ and $\mathcal{I}$ based on algorithm~\ref{selection}. Based on the minimum angle $\alpha$ for $p^{f} \in \mathcal{D}$ and minimum cost $\mathbb{C}$ for $p^{f} \in \mathcal{I}$ a candidate frontier $p^{f}_{candidate}$ is selected. As presented in Figure ~\ref{fig:autonomy}, $D^{*}_{+}$ takes the $p^{f}_{candidate}$ and plans a risk-aware path to $p^{f}_{candidate}$. All consecutive points from the path planned by $D^{*}_{+}$ are sent to APF (Artificial Potential Fields) module which shift the way point if it lies close to an occupied point resembling the low level obstacle avoidance incorporated within the planning step. The shifted way point is sent to the nMPC controller which is responsible for actuating the MCQ in order to move to the first point in path. The loop runs at 20 $Hz$ while the exploration is in progress. Detailed explanation of the planing-control-avoidance architecture is presented in subsections ~\ref{subsec:control},\ref{subsec:avoidance} and \ref{subsec:planning}.

\subsubsection{Nonlinear Model Predictive Control of the Coaxial Quadrotor}\label{subsec:control}
While there are many different control approaches to a MAV system, we will utilize a  NMPC for reference tracking and stable flight behavior, as successfully utilized in previous applications for MAVs as in~\citep*{lindqvist2020nonlinear}, and can be found in \autoref{eq:mavkinematic}. This model considers eight states in a yaw-compensated body-frame namely: positions coordinates $p^\mathcal{B}=[p_x^\mathcal{B},p_y^\mathcal{B},p_z^\mathcal{B}]^\top$, linear velocities $v^\mathcal{B} = [v_x^\mathcal{B},v_y^\mathcal{B},v_z^\mathcal{B}]^\top$ as well as roll and pitch angles $\phi$ and $\theta \in[-\pi,\pi]$, and computes control inputs as $T, \phi_{\mathrm{ref}}, \theta_{\mathrm{ref}}\in \mathbb{R}$ where $T \geq 0$ is the total mass-less thrust produced by the motors. Furthermore, the model simplifies the MCQ dynamics by assuming the existence of a low-level attitude controller and modelling the roll/pitch angles as first-order systems (let $K_{\phi}, K_{\theta}\in \mathbb{R}$ and $\tau_{\phi}, \tau_{\theta} \in \mathbb{R}$ denote the gains and time constants) as to approximate the closed-loop behavior of the attitude controller (here utilizing the the RotorS framework modified to fit the co-axial Quadrotor \citep*{Furrer2016}). Based on this model, let us define the state vector as $x = [p^\mathcal{B}, v^\mathcal{B}, \phi, \theta]^\top$, and denote the control inputs as: $u=[T,\phi_{\mathrm{ref}},\theta_{\mathrm{ref}}]^\top$. Thus, the dynamic model of the MCQ can be provided as:
\begin{subequations}
\label{eq:mavkinematic}
\begin{align}
        &\dot{p}^\mathcal{B}(t)  = v^\mathcal{B}(t) \\ 
        &\dot{v}^\mathcal{B}(t)  = R(\phi,\theta) 
        \smallmat{0 \\ 0 \\ T} + 
        \smallmat{0 \\ 0 \\ -g} - 
        \smallmat{A_x & 0 & 0 \\ 0 &  A_y & 0 \\ 0 & 0 & A_z} v^\mathcal{B}(t), \\ 
        &\dot{\phi}(t)  = \nicefrac{1}{\tau_{\phi}} (K_\phi\phi_{\mathrm{ref}}(t)-\phi(t)), \\ 
        &\dot{\theta}(t)  = \nicefrac{1}{\tau_{\theta}} (K_\theta\theta_{\mathrm{ref}}(t)-\theta(t)).
\end{align}
\end{subequations}
The model is then discretized by the forward Euler method to achieve the predictive form $    x_{k+1} = \zeta(x_k, u_k)$.

%

The discrete model is used as the prediction model of the NMPC and the prediction horizon is denoted by $N$. A cost is associated to the states and control inputs at current time and using nonlinear optimizer, optimal set of control actions can be optimized corresponding to the minimum cost. Let $x_{k+i|k}$ represent the prediction state at time step $k+i$, produced at time step $k$. The corresponding control action can be denoted as $u_{k+i|k}$. Therefore, $\textbf{x}_{k}$ and $\textbf{u}_{k}$ can be denoted as the full predicted states and control inputs. The cost function~\citep*{lindqvist2020nonlinear} is defined as: 
\begin{align*}
    \mathcal{C}(\textbf{x}_{k},\textbf{u}_{k},u_{k-1|k}) =  \sum_{i=0}^{N} \overbrace{\mathcal{W}_{x}{||x_{ref}-x_{k+i|k}||}^2}^\text{state cost} + \\
    \overbrace{\mathcal{W}_{u}{||u_{ref}-u_{k+i|k}||}^2}^\text{input cost} + \overbrace{\mathcal{W}_{\Delta u}{||u_{k+i|k}-u_{k+i-1|k}||}^2}^\text{Input rate cost}
\end{align*}
Where, $\mathcal{W}_{x} \in \mathbb{R}^{8\times8}$ and $\{\mathcal{W}_{u}, \mathcal{W}_{\Delta u}\} \in \mathbb{R}^{3\times3}$ are adaptive weight matrices for the states, control inputs and input rates respectively. In the above mentioned cost function, the state cost penalizes the deviation from the reference trajectory, the input cost penalizes the reference input needed for hovering and the input rate penalizes the control input rates to provide a smooth trajectory to reach the desired state of the system. The input rate cost allows the MCQ to avoid aggressive control actions, which promotes stable operation while minimizing the change in attitude of the MCQ. Additionally, we impose constraints on the inputs and input-rates respectively.
The constraints on control inputs are defined as $u_{min} \leq u_{k+i|k} \leq u_{max}$ and the constraints on rate of control input are defined as:
\begin{align}
    | \phi_{(ref,k+i-1|k)} - \phi_{(ref,k+i|k)} | \leq \Delta \phi_{max}  \\
    | \theta_{(ref,k+i-1|k)} - \theta_{(ref,k+i|k)} | \leq \Delta \theta_{max} 
    \label{rateconst}
\end{align}
In Equation~\ref{rateconst}, $\Delta \phi_{max}$ and $\Delta \theta_{max}$ represent the maximum allowed rate of change in roll and pitch for each time step. In order to minimize the aforementioned cost function, the optimization problem is solved using the Optimization Engine \citep*{sopasakis2020open} where a penalty method is used to compute solutions that satisfy the input rate constraints. 

\subsubsection{Collision avoidance}\label{subsec:avoidance}

While obstacle-free paths are provided by the $D^{*}_{+}$ module, the frontier based exploration framework should still have decent level of local obstacle avoidance in order to maintain a safe distance to obstacles and walls in the environment, or as an additional safety layer in the case of error in occupancy information within planning module. This work utilizes a purely reactive 3D Artificial Potential Field (APF), described in more detail in~\citep*{lindqvist2021compra}, relying only on the instantaneous pointcloud stream from the 3D lidar. It is based on letting each point in the lidar pointcloud within a specified volume defined by radius $r_F$ result in a repulsive force, and then summing all such point-forces to get the total repulsive force. Let us denote the local point cloud generated by the 3D lidar as $\{P\}$, where all points are described by a relative position to the lidar as $\rho = [\rho_x, \rho_y, \rho_z]$. Also denote the repulsive force as $F^r = [F_{x}^r, F_{y}^r, F_{z}^r]$. Denote the list of points $\boldsymbol{\rho}_r \in \{P\}$, where $\mid\mid \rho_r^i \mid \mid \leq r_F$ and $i = 0,1, \ldots, N_{\rho_r}$ (and as such $N_{\rho_r}$ is the number of points to be considered for the repulsive force). The repulsive force is $F^r = \sum^{N_{\rho_r}}_{i=0}  L(1 -  \frac{\mid\mid \rho_r^i \mid\mid}{r_F})^2\frac{-\rho_r^i }{\mid\mid \rho_r^i \mid\mid}$
where $L$ is the repulsive constant. The attractive force $F^a$ can be seen as the vector from $\hat{p}^\mathcal{B}$ to the next way-point provided by D$^*$-lite as $F^a = w_1^{\mathcal{B}} - \hat{p}^{\mathcal{B}}$. Additionally, we impose saturation limits on the magnitude of forces, the rate of change of forces and also normalizing the resulting total force $F$ as to always generate forces of the same magnitude and to enforce stable and non-oscillatory flight behavior. The output is the obstacle-free position reference given to the NMPC as $p_{ref}^{\mathcal{B}} = F + \hat{p}^{\mathcal{B}}$, where $p_{ref}^{\mathcal{B}}$ are the first three elements of $x_{\mathrm{ref}}$, such that $x_{\mathrm{ref}} = [p_{ref,x}^{\mathcal{B}}, p_{ref,y}^{\mathcal{B}},  p_{ref,z}^{\mathcal{B}}, 0, 0, 0, 0, 0]^\top$. 

\subsection{Incrementally Built Map Based Planning}\label{subsec:planning}

\begin{figure}[h!]
  \centering
    \includegraphics[width=\linewidth]{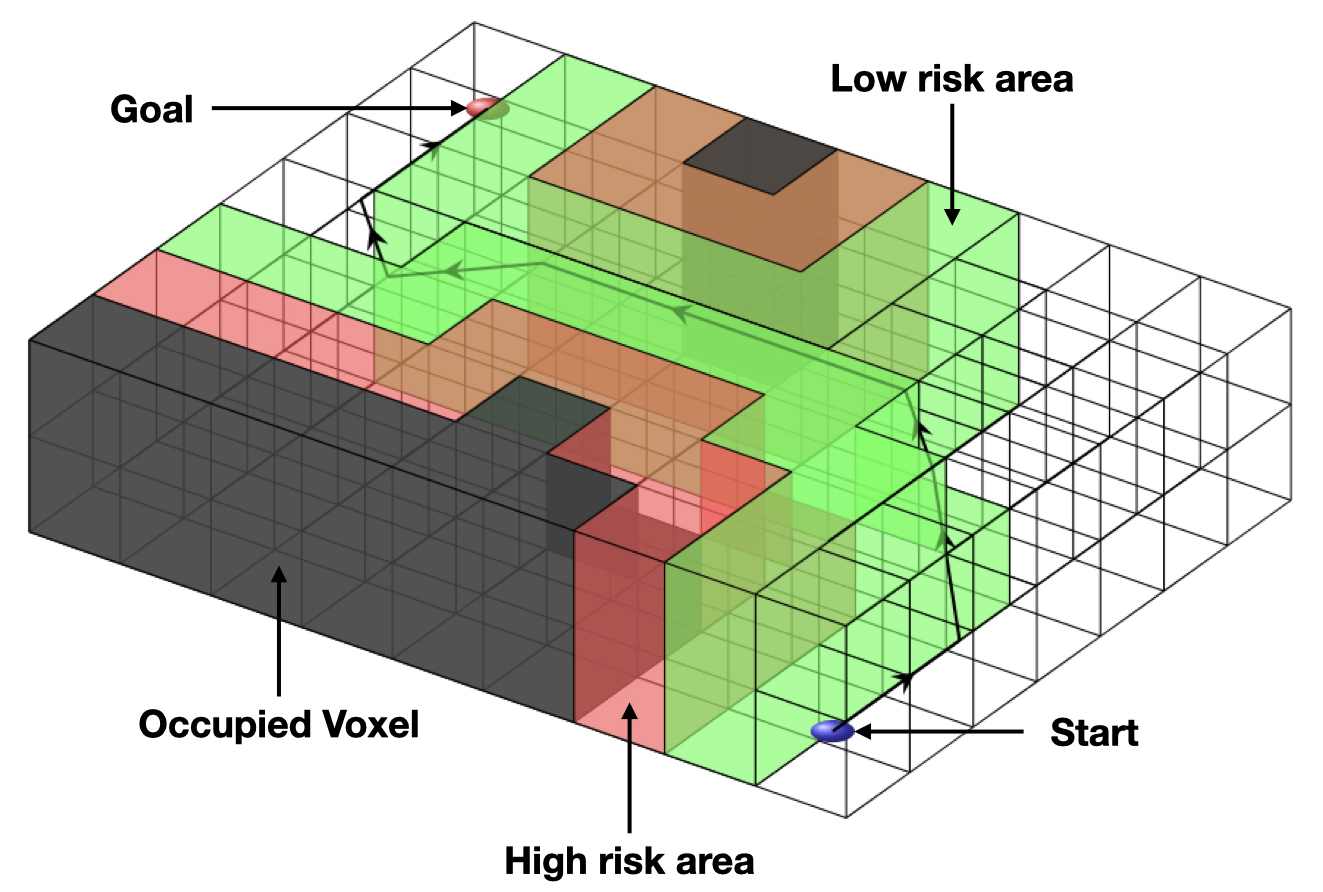}
    \caption{A illustration of how DSP plans a path from the blue ball to the red ball with regards to the risk layer, red voxels is higher risk, green voxels is lower risk, and the black voxels are occupied.}
  \label{fig:riskViz}
\end{figure}

The path planning module is based on a modified version of D$^*$-lite~\citep*{dslgridsearch} called D$^*_+$ (DSP), and the occupancy mapper Octomap~\citep*{hornung2013octomap}. The modification to D$^*$-lite is to its internal map representation ($\mathbb{G}$) where a representation for unknown areas is added as well as a risk-cost associated to space close to occupied spaces. By assigning different traversal costs $C$ to the risky areas, DSP is made to plan the safest and shortest path ($P$) to the destination, such that $P$ is the path in $\mathbb{G}$ where $\sum{\forall C \in P}$ is minimized. Here DSP is used to plan the map-based obstacle-free path from the agents current position to the position of the selected frontier.
$C$ is assigned to each Voxel sow that:
$$
C_n=\begin{cases}
                    1, & \text{if $Free$}\\
                    C_o, & \text{if $Occupied$} \\
                    C_u, & \text{if $Unknown$} \\
                    C_n + \frac{c_u}{d+1}, & \text{if $d < r$ (only added once per Voxel)} 
                \end{cases}
$$  
where $d$ is the distance in Voxels to the closes occupied Voxel, and $r$ is the range that are consider risky from a occupied Voxel measured in Voxels.
Values of $C_o$ should be set to a significant higher value than $C_u$ is set to.
$C_u$ can in turn be set to a low value to encourage some exploration during traversal of the path or to a higher to have safer paths within known free space. 
In \autoref{fig:riskViz} is an example path of how DSP plans $P$ in free space where it is possible and only enters a risky area when there is no other option.
This behavior is achieved with the gradient risk for proximity to occupied Voxels.
A frontier Voxel is marked inaccessible when the size of the MCQ is higher than the distance between two occupied Voxels through which a safe path to a frontier can be planned. Such formulation is advantageous in discarding frontiers which require extremely risky path to traverse to the point. The detailed study in risk aware occupancy information based path planning is presented in our previous work~\citep*{karlsson2021d}.

%
\section{Lava Tube Simulated Exploration Mission}\label{sec:sim_results}
\begin{figure}[h!]
  \centering
    \includegraphics[width=\linewidth]{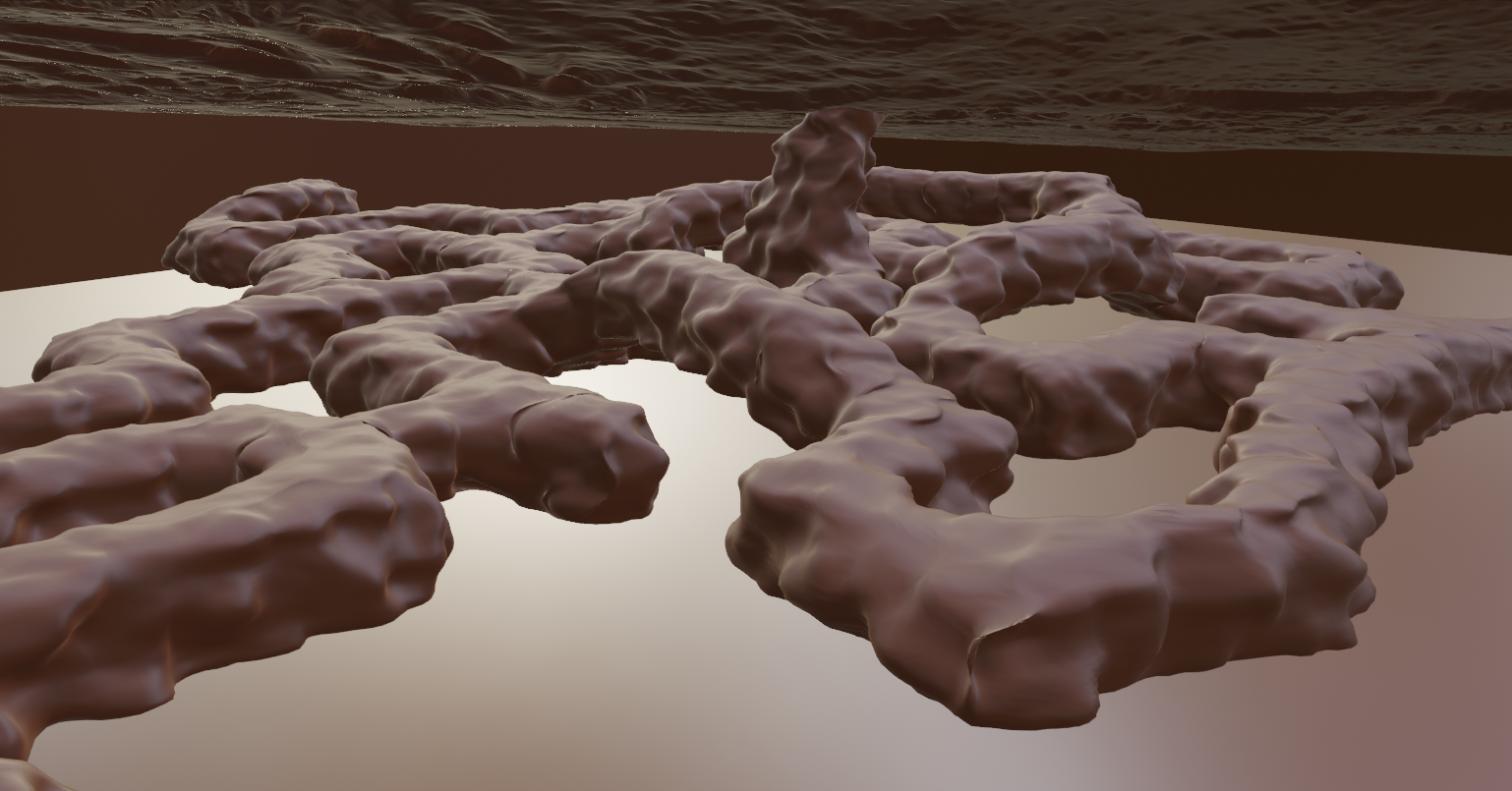}
    {\caption{Open sourced realistic lava tube environment created and used for the Gazebo simulations.}}
  \label{fig:lavatube}
\end{figure}

\begin{figure}[h!]
    \centering
    \subfigure[]
    {
        \includegraphics[width=0.476\linewidth]{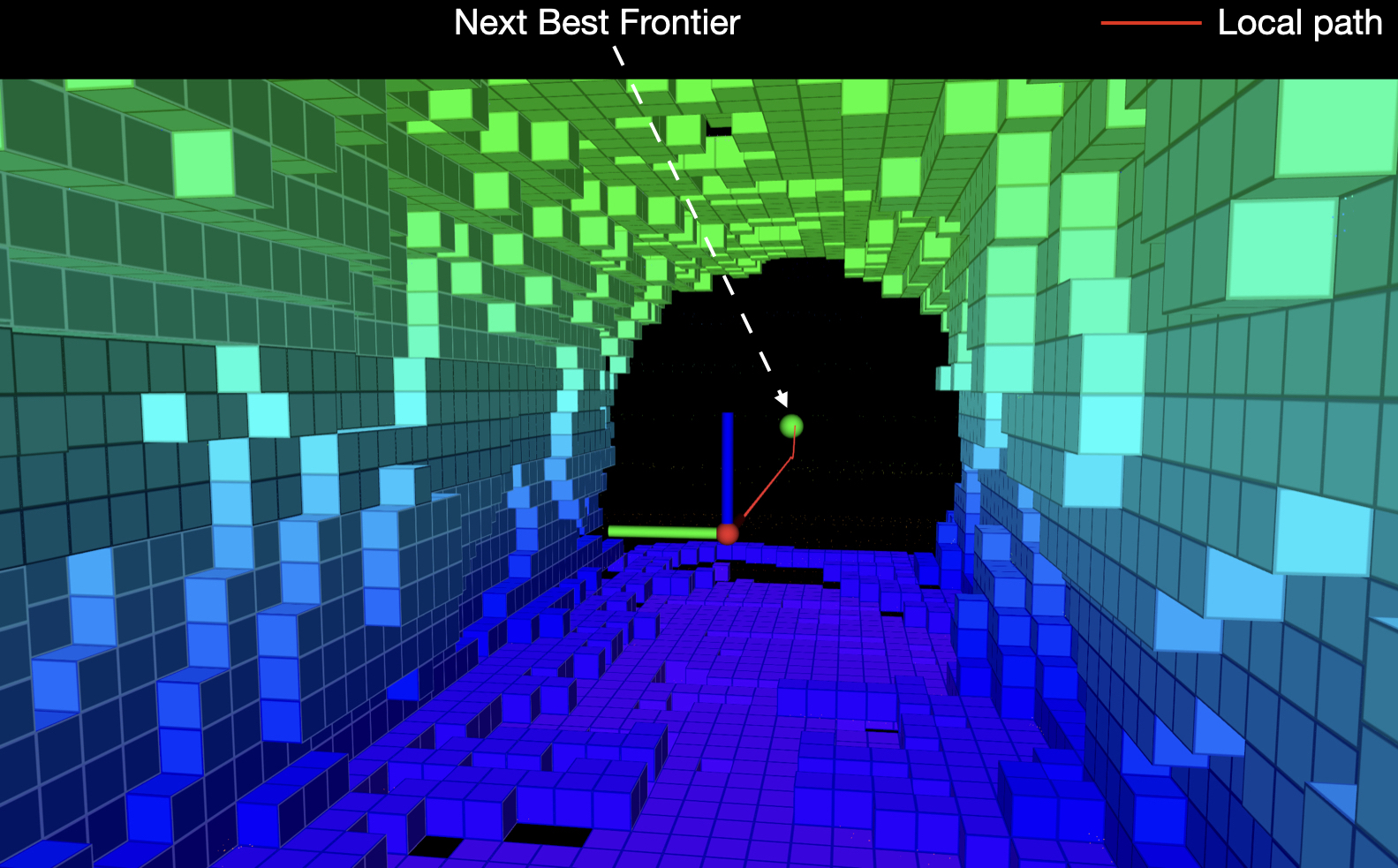}
        
    }
    \subfigure[]
    {
        \includegraphics[width=0.464\linewidth]{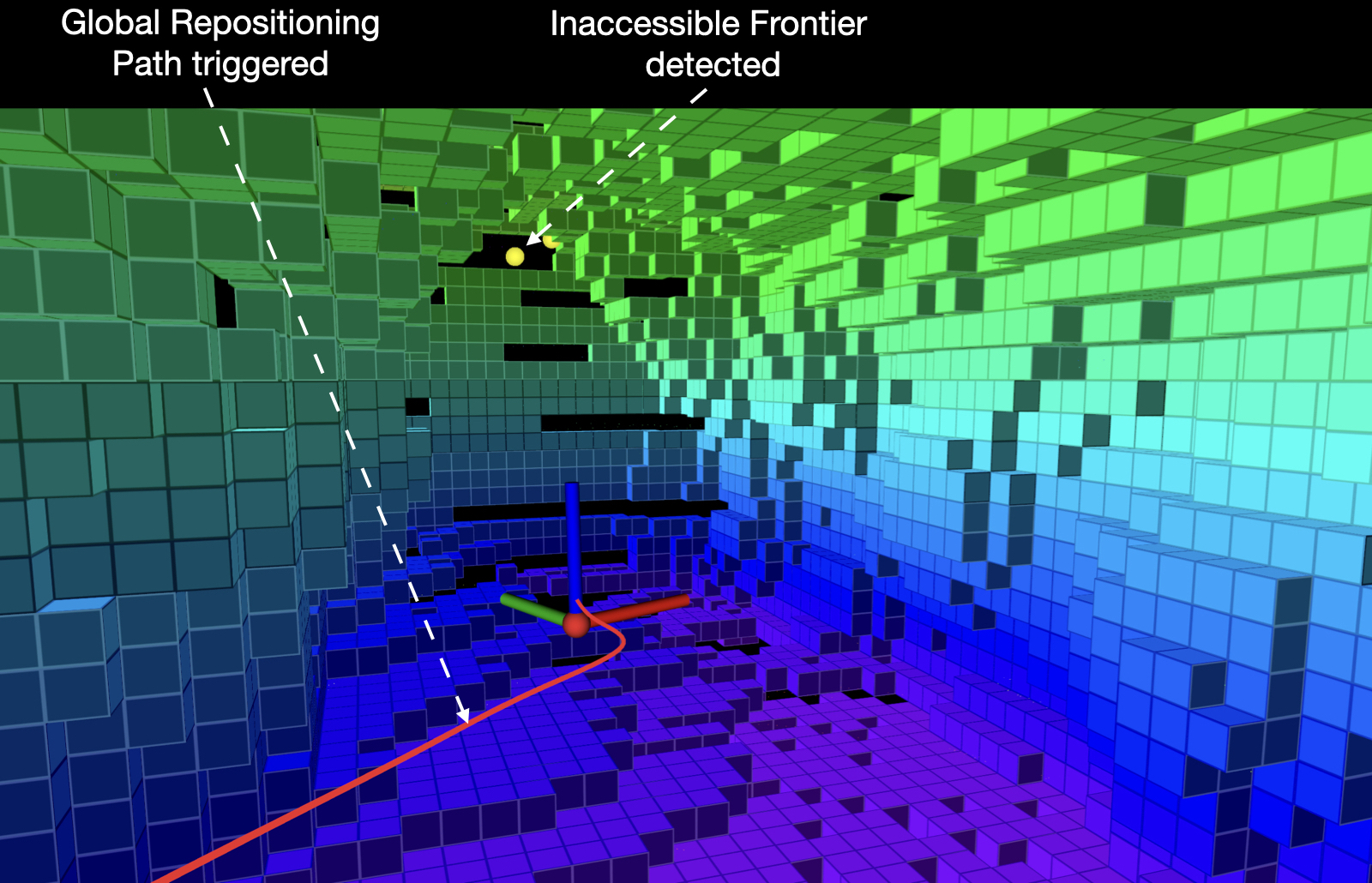}
        
    }
    {\caption{(a) Next Best frontier with respect to MCQ's current position (b) Inaccessible Frontier detection and and triggered homing maneuver.}}
    \label{fig:nbf}
\end{figure}

\begin{figure}[h!]
  \centering
    \includegraphics[width=\linewidth]{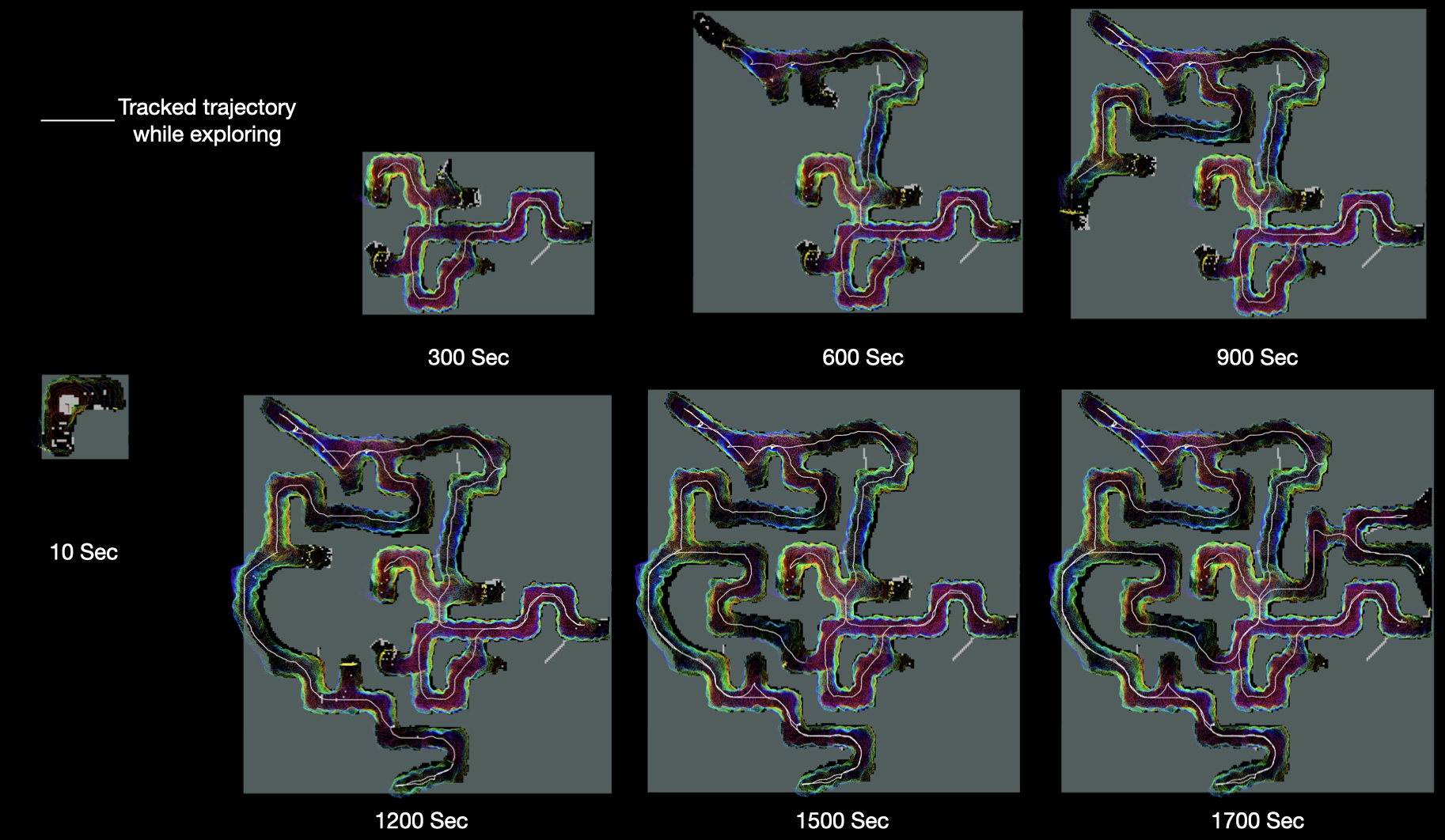}
    {\caption{Incremental exploration of Mars lava tube using proposed autonomy architecture.}}
  \label{fig:incremental}
\end{figure}

\begin{figure}[h!]
  \centering
    \includegraphics[width=\linewidth]{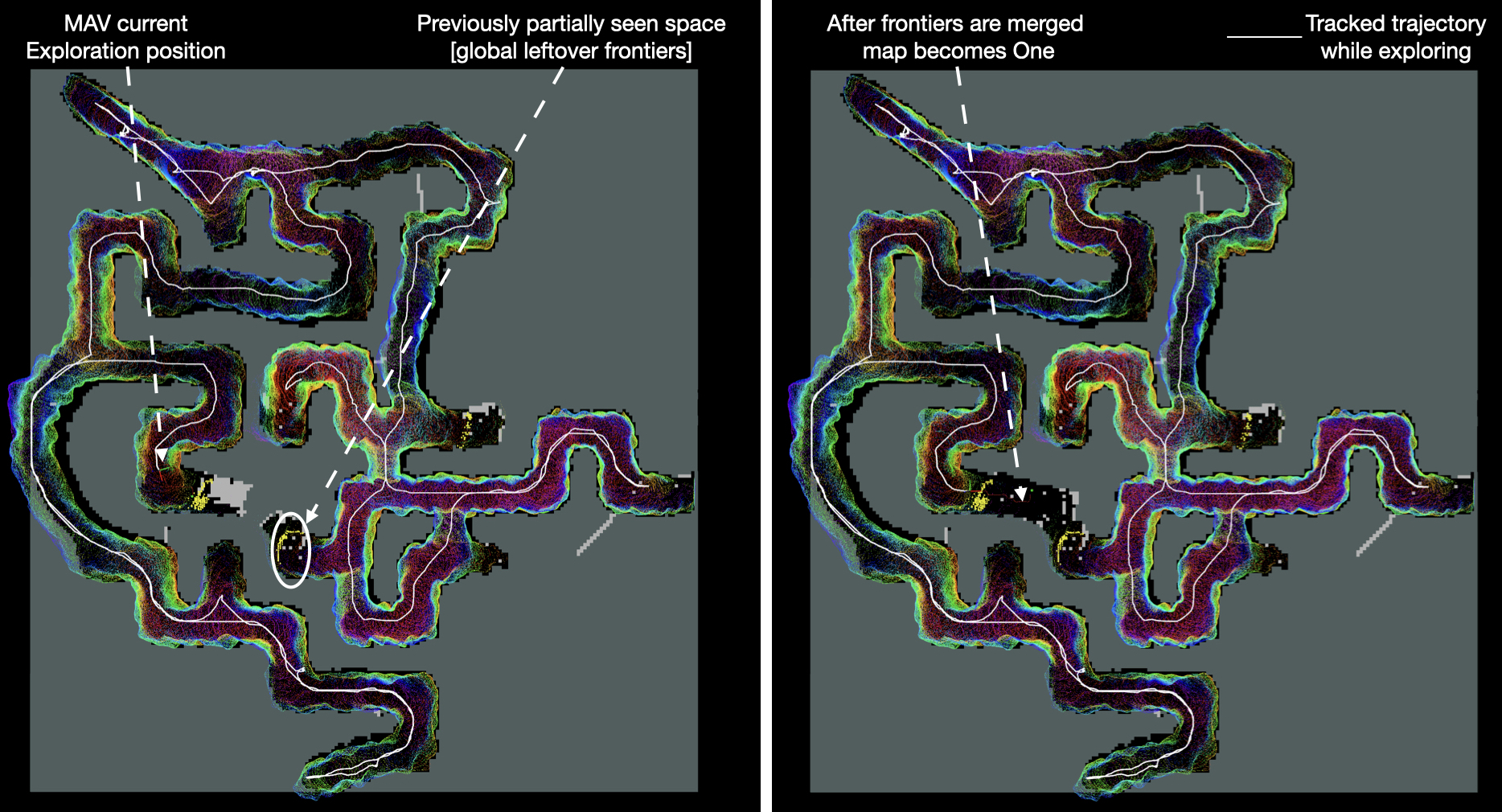}
    \caption{Global frontiers are merged when previously partially seen area is discovered again.}
  \label{fig:mapstiched}
\end{figure}

A gazebo world is constructed to replicate true potential Mars lava tube environment used in the simulated exploration mission in this work as presented in \autoref{fig:lavatube}. We have made the Mars lava tube gazebo world available for public use at: \url{https://github.com/LTU-RAI/MarsLavaTubeWorld.git}.
The simulations are carried out using a true Mars atmospheric model in gazebo in order to simulate exploration mission with utmost realistic scenario corresponding to surface pressure and gravity. As described in \autoref{sec:coax_quad}, we use Mars Coaxial Quadrotor with a top mounted VLP-16 Lidar which has $30^\circ$ vertical FOV and $360^\circ$ horizontal FOV. The simulations are carried out on a Intel core i7 unit, with 32GB RAM. We use Rotor Simulator \citep*{Furrer2016} with a custom model of Mars Coaxial Quadrotor as a simulation framework. All visualization representations are derived from Rviz. As presented in \autoref{sec:IFE}, the proposed framework computes for safe next best frontier in the field of view of the MCQ and the corresponding local planning framework plans an obstacle free path to such frontier pose. The same can be visualized in \autoref{fig:nbf}. The green marker represents the candidate frontier while the red path is the safe local path to such $p^{candidate}$. When there exist no $safe$ directly accessible frontier ahead of the MCQ, an indirectly accessible frontier is computed based on the cost function described earlier in \autoref{sec:IFE}. This also applies when inaccessible frontiers are generated due to the uncertainty in octree formation or simply because of the holes in Octomap. A global re-positioning is triggered in such cases to re locate the MCQ to previously partially seen areas of the lava tube. The same behaviour can be visualized in \autoref{fig:nbf}. 
The full (99$\%$) exploration of the Virtual Mars lava tube is achieved by the proposed framework in a 1700 second exploration mission. As it is in all exploration mission, at the beginning of exploration the MCQ has $zero$ information available about the environment. Using the proposed Incremental Frontier Exploration approach with a global planing and collision avoidance modules, the MCQ explores the lava tube environment. in \autoref{fig:incremental}, a snap shots of incremental map building is presented with the interval of 300 seconds. In \autoref{fig:incremental}, at 10 seconds into the mission, the MCQ has just started exploring the environment while at 1700 seconds the MCQ has almost (99$\%$) explored the full lava tube. 
Due to the complexity of the lava tube environment, there exist some parts of the tube where the MCQ meets dead ends while exploring locally. At such points it is essential that the MCQ does not spend more time hovering around a dead end and autonomously re-position itself to a different part of the tube where more information gain is believed. The proposed framework emphasise on global re-positioning of the MCQ in such scenarios by computing a frontier from $\mathcal{I}$ based on the traverse cost as well as actuation cost with respect to the current location of the MCQ and potential global re positioning poses. This behaviour is clearly visualized in \autoref{fig:localandglobal} in which as soon as the MCQ meets a dead-end/inaccessible frontier and a global re-positioning path (green) is computed which will re locate the MCQ to a previously partially seen (while exploring) part of the lava tube. The proposed framework also keeps track of globally leftover frontiers and at some point in the exploration mission if the global leftover frontiers (previously partially seen areas) as found in the field of view and if such frontiers are accessible then they are merged with the current $\mathcal{D}$ frontiers set and the map is merged at such points. This behaviour is presented in visualization form in \autoref{fig:mapstiched}. In such scenario after merging the frontiers, the MCQ re positions itself to another part of the tube where it continues exploring the tube so that rapid, energy efficient yet safe exploration mission behaviour can be maintained.  

\begin{figure}[h!]
  \centering
    \includegraphics[width=\linewidth]{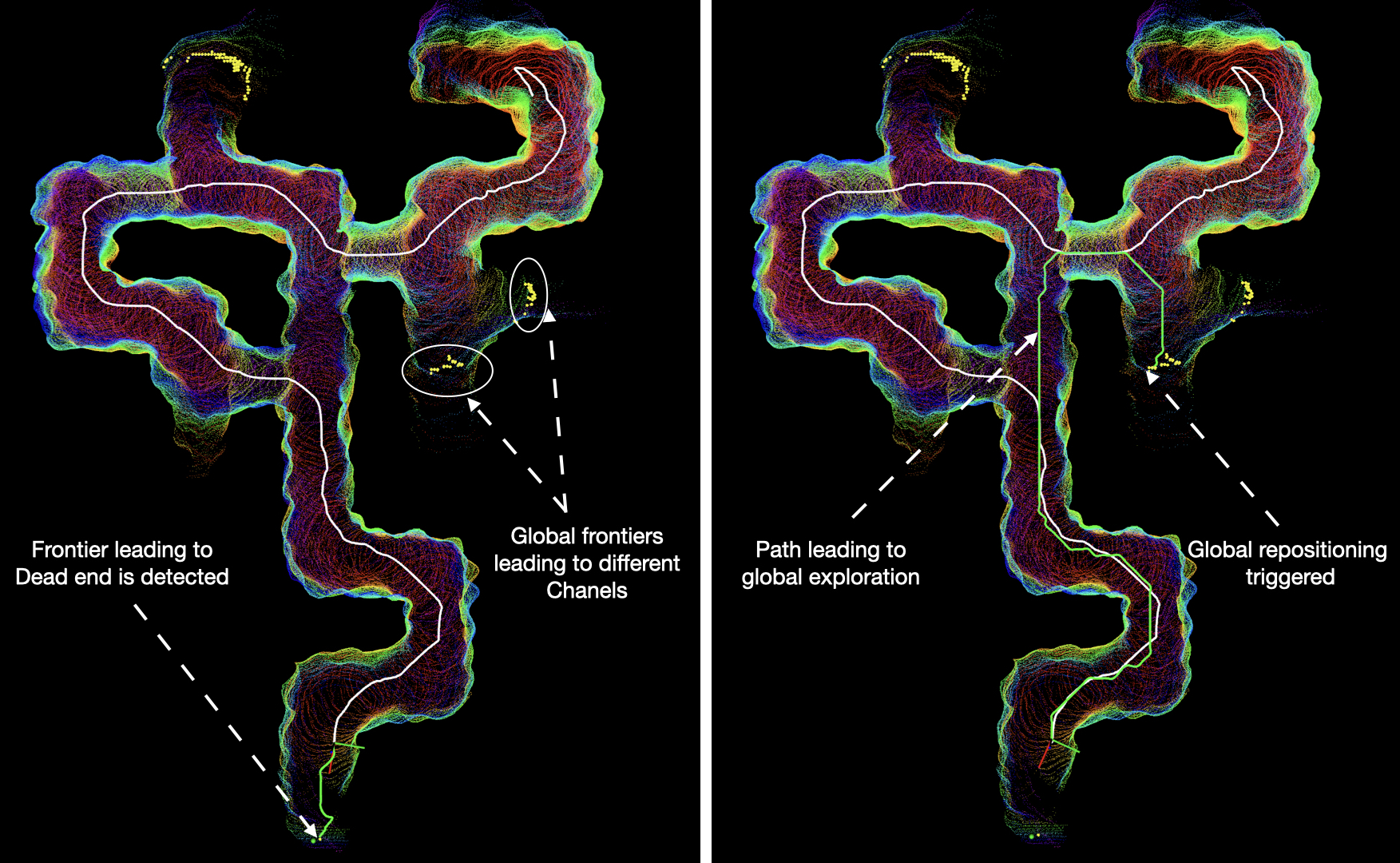}
    {\caption{Global re-positioning due to inaccessible frontier (dead end) ahead of the MCQ.}}
  \label{fig:globalrepos}
\end{figure}

\begin{figure}[h!]
  \centering
    \includegraphics[width=\linewidth]{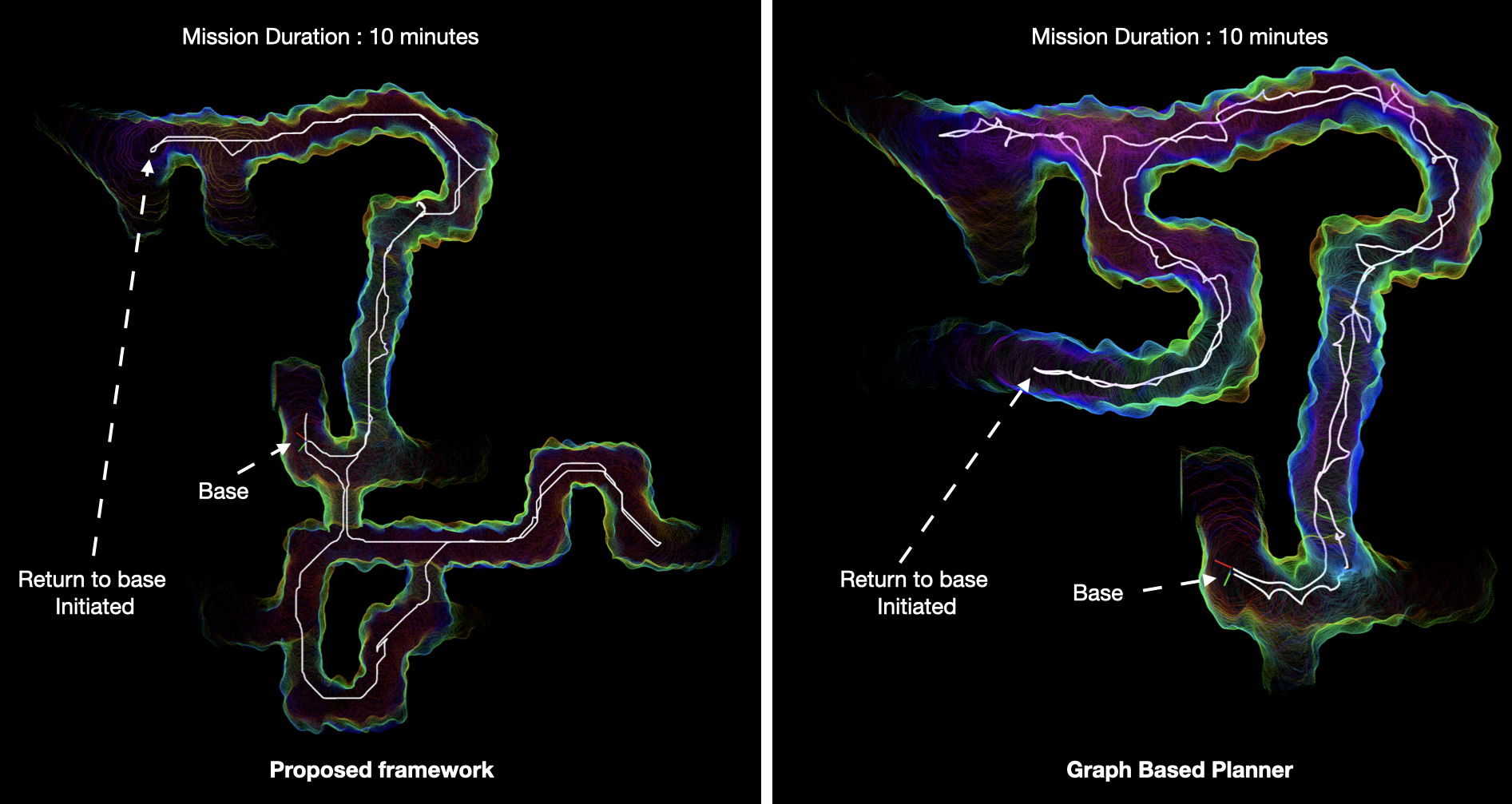}
    \caption{Mission duration : 10 min | Mars lava tube exploration comparison against Graph Based Planner2.}
  \label{fig:10minmission}
\end{figure}

\begin{figure}[h!]
  \centering
    \includegraphics[width=\linewidth]{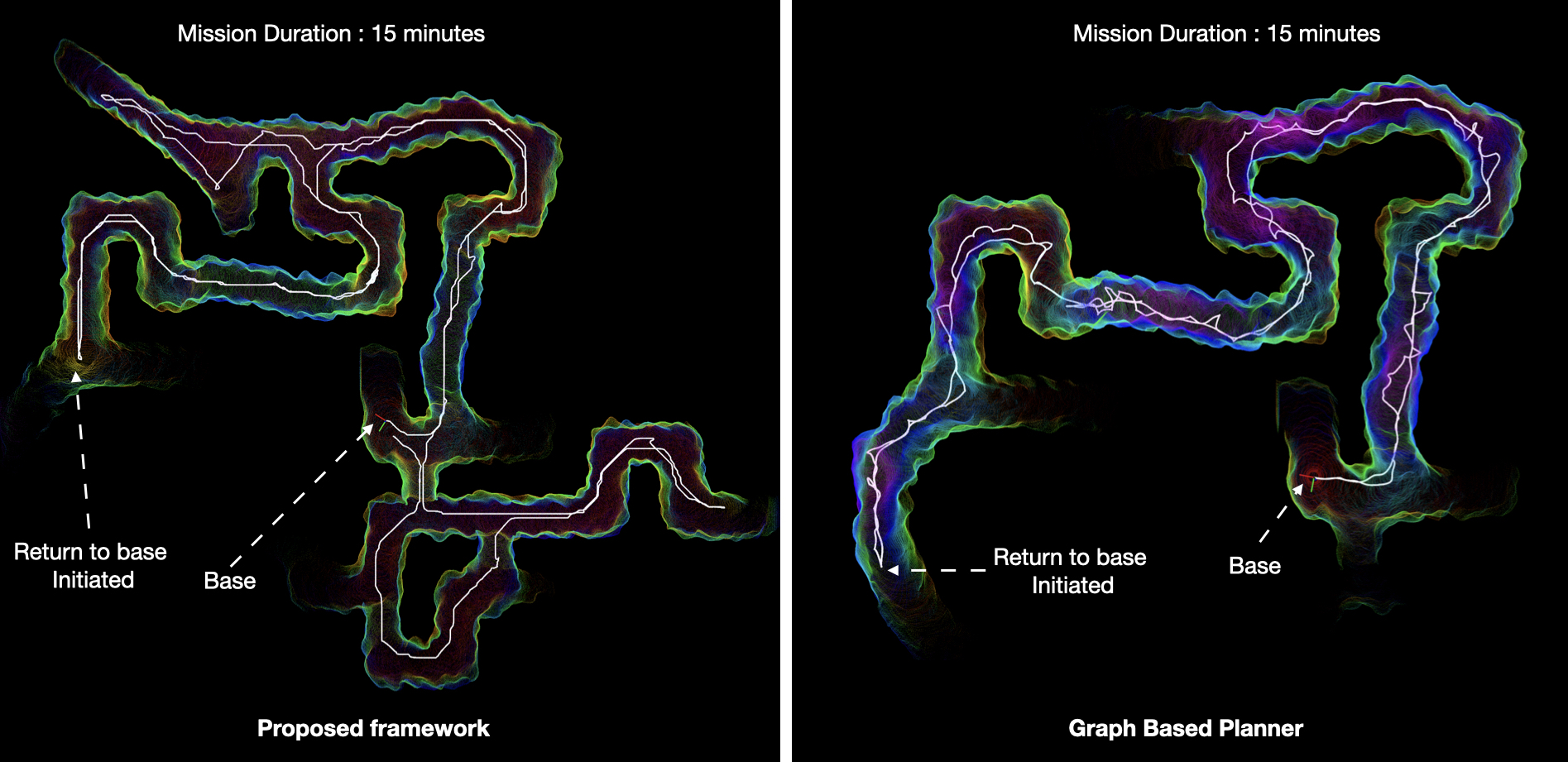}
    \caption{Mission duration : 15 min | Mars lava tube exploration comparison against Graph Based Planner2.}
  \label{fig:15minmission}
\end{figure}

In order to validate and benchmark the proposed rapid yet safe exploration framework, we also compare our exploration results with the State-of-the-Art Graph Based Planner~\citep*{dang2020graph} which is revised and modified to use in real subterranean worlds by the winning team at DARPA Sub-T Challenge~\citep*{kulkarni2021autonomous}. In order to make fair comparison we simulate multiple Mars lava tube exploration mission using the both methods with exactly the same parameters in terms of maximum forward velocity, sensor parameters, mission time, simulation set up etc. Since the proposed work is addresses an utmost autonomous exploration of Mars lava tube, we cap the exploration time limit at 10 minutes and 15 minutes in two exploration scenarios for both methods. This relates to a realistic MCQ flight duration and it yields real exploration volume in constrained mission time. For both methods, autonomous home return is triggered as soon as times exceeds the mission duration (10 and 15 minutes). In \autoref{fig:10minmission}, the part of the explored lava tube by the two methods is shown where the mission duration is 10 minutes. The MCQ start and return at the same position in both methods however, the path taken by the MCQ differs due to the computed information gain or directly accessible frontier by the two method differs. The similar mission is also carried out and the MCQ trajectory and the explored lava tube is presented in \autoref{fig:15minmission}. It is evident that the using the proposed method the MCQ covers more ground and also explores more unexplored part of the lava tube in the same time as compared to the graph based planner. In \autoref{fig:10minmission} and \autoref{fig:15minmission} it is also evident that the MCQ trajectory tracked is also different due to the fact that the proposed method implements a risk aware global planning method to next best view points where as the graph based planner plans a safe trajectory to maximum information gain nodes which often leaves the MCQ to go back and forth before the global planner is triggered when local information gain is low. 

\begin{figure}[h!]
    \centering
    \subfigure[]
    {
        \includegraphics[width=0.47\linewidth]{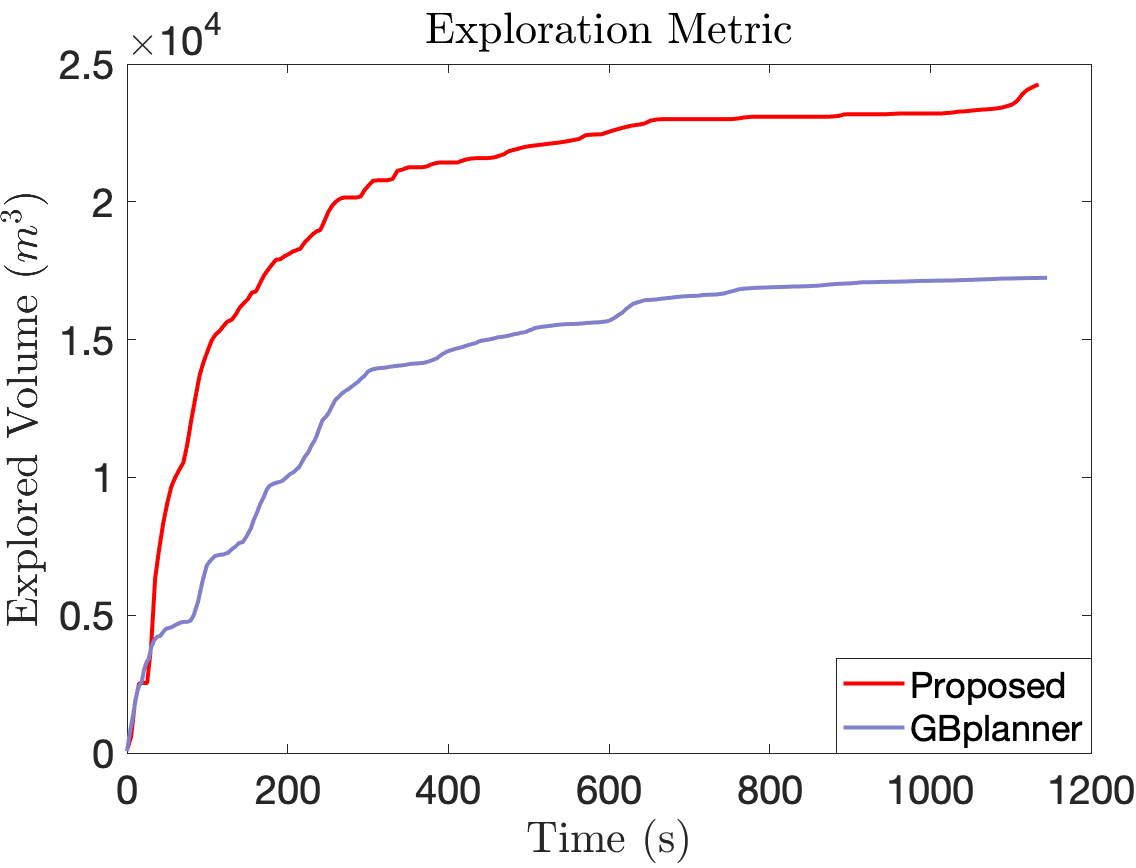}
        
    }
    \subfigure[]
    {
        \includegraphics[width=0.47\linewidth]{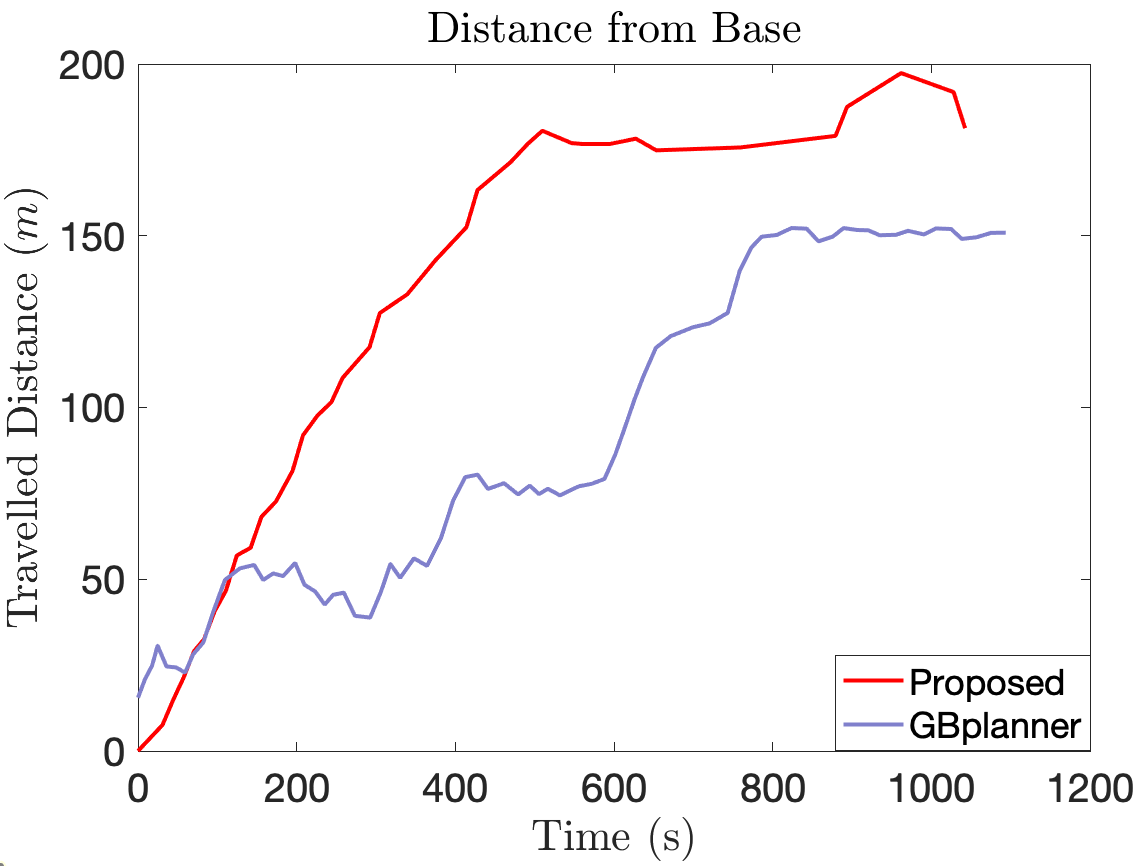}
        
    }
    \caption{Exploration mission comparisons in a non fixed time budget mission. (a) Explored volume (b) Traveled distance from the base.}
    \label{fig:fullmissioncompare}
\end{figure}

\begin{figure*}[h!]
    \centering
        \includegraphics[width=\linewidth]{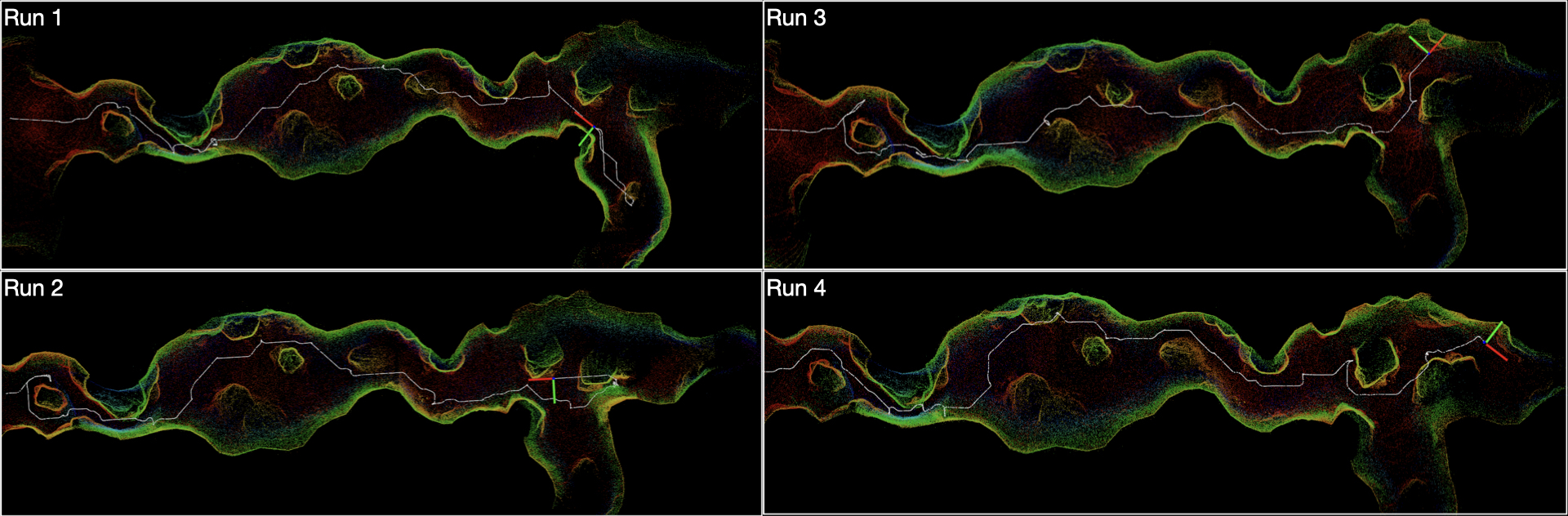}
    {\caption{The explored parts of lava tube and different paths planned by the MCQ in four different exploration runs.}}
    \label{fig:fourruns}
\end{figure*}

\begin{figure}[h!]
    \centering
    \subfigure[]
    {
        \includegraphics[width=0.47\linewidth]{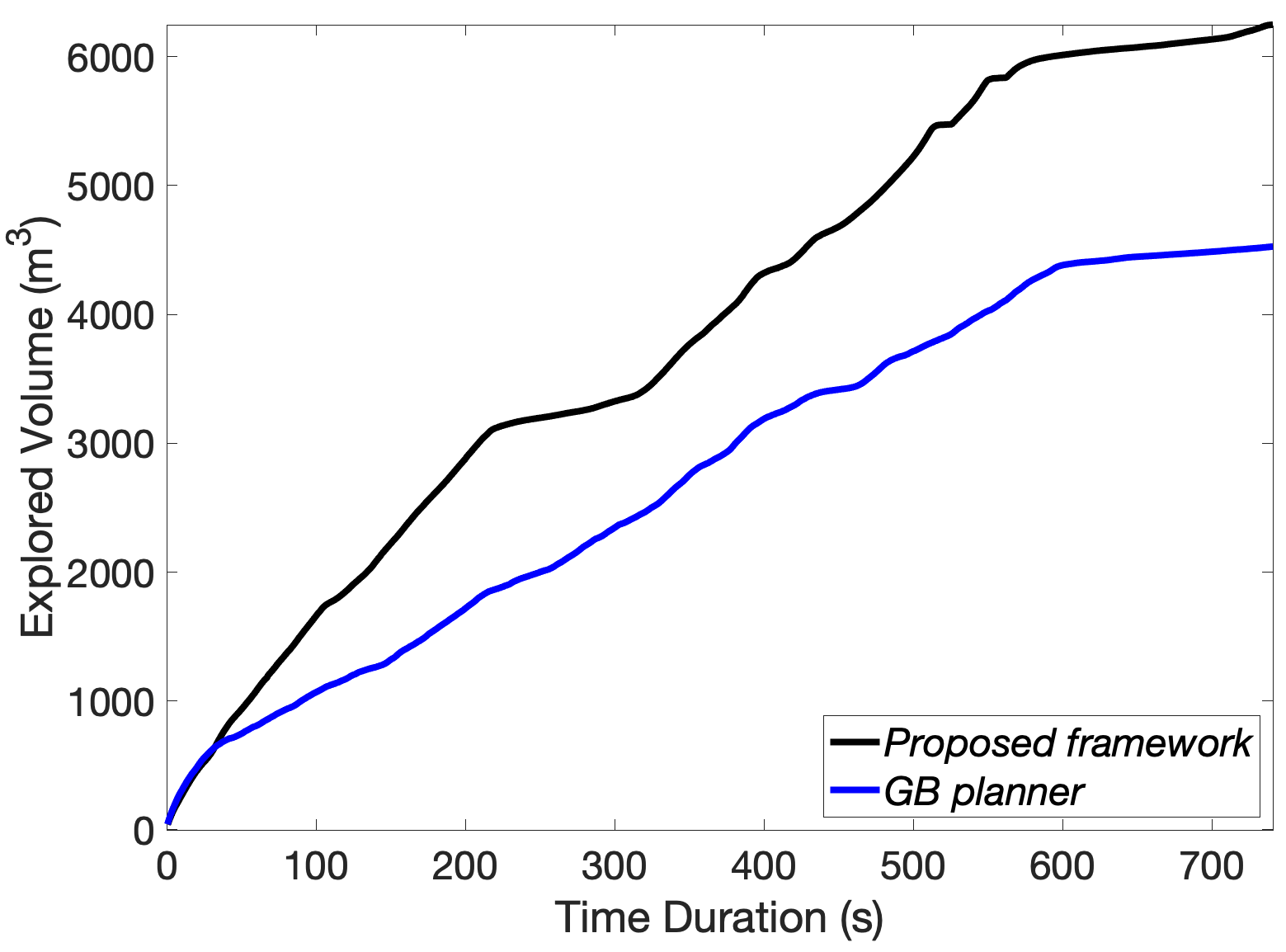}
        
    }
    \subfigure[]
    {
        \includegraphics[width=0.47\linewidth]{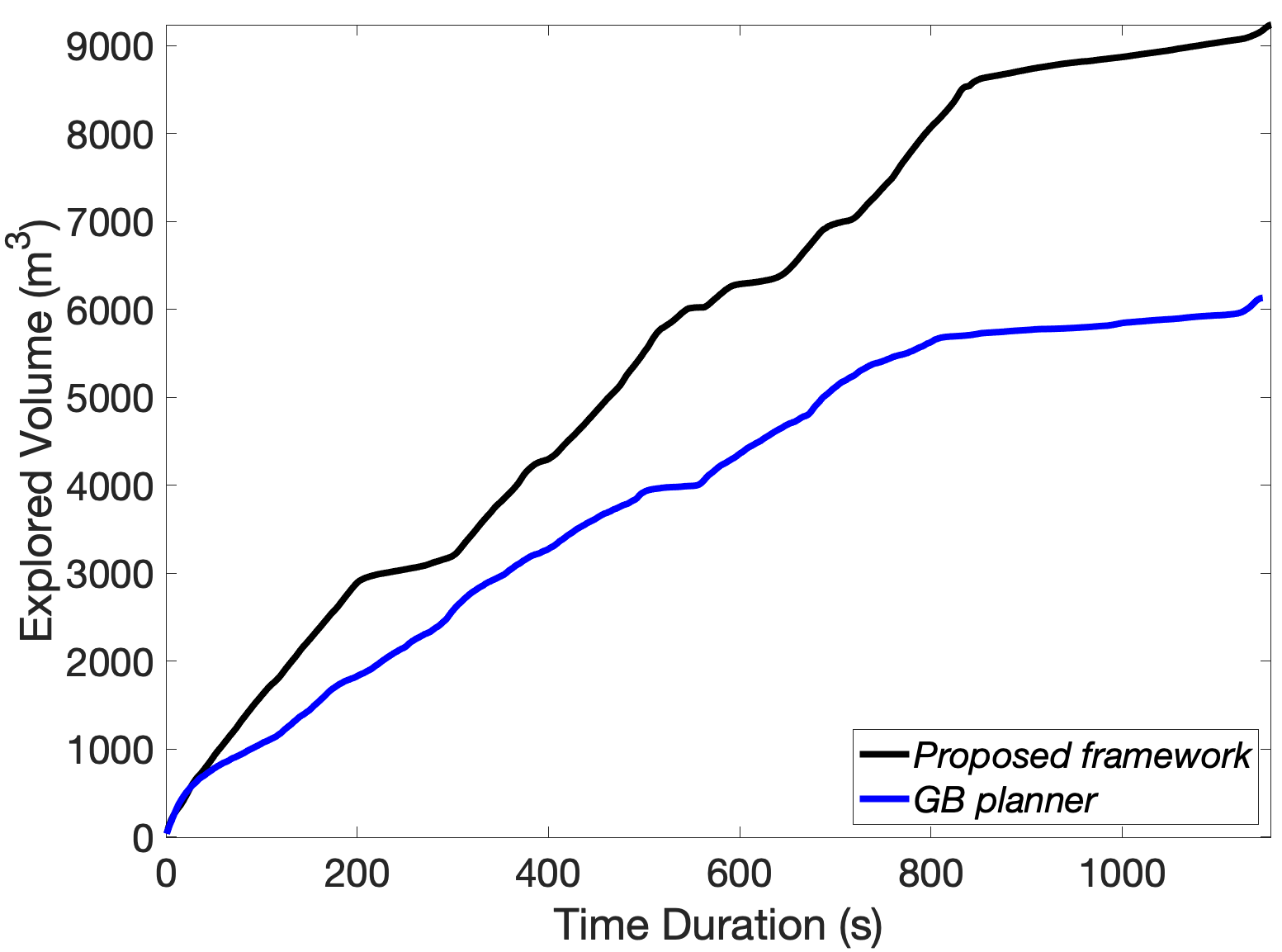}
        
    }
    {\caption{Exploration Volume gain for fixed-time budget based missions (a) 10 minute exploration mission (b) 15 minute exploration mission.}}
    \label{fig:volumeexplored}
\end{figure}

\begin{figure}[h!]
    \centering
    \subfigure[]
    {
        \includegraphics[width=0.47\linewidth]{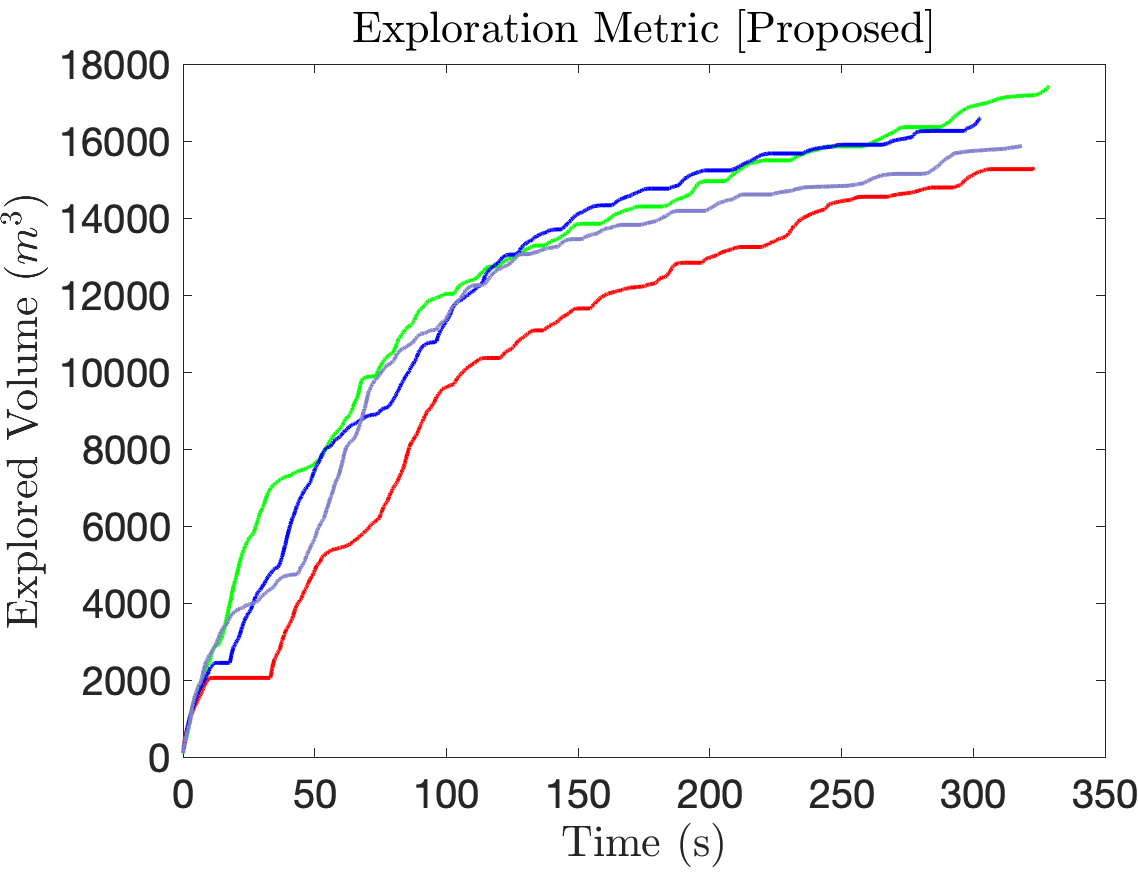}
        
    }
    \subfigure[]
    {
        \includegraphics[width=0.47\linewidth]{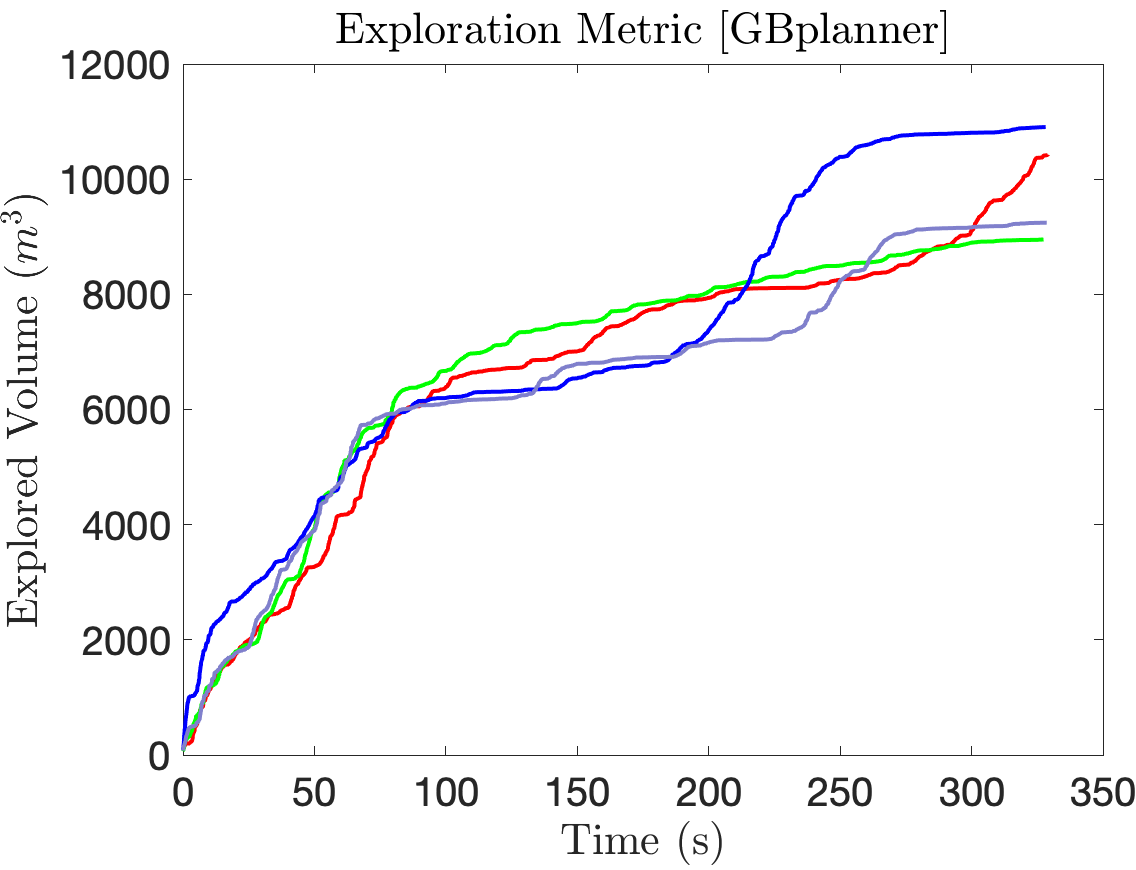}
        
    }
    \caption{Exploration Volume gain for 4 runs (a) Proposed approach (b) GB planner.}
    \label{fig:volumeexplorednew}
\end{figure}

\begin{figure}[h!]
    \centering
    \subfigure[]
    {
        \includegraphics[width=0.47\linewidth]{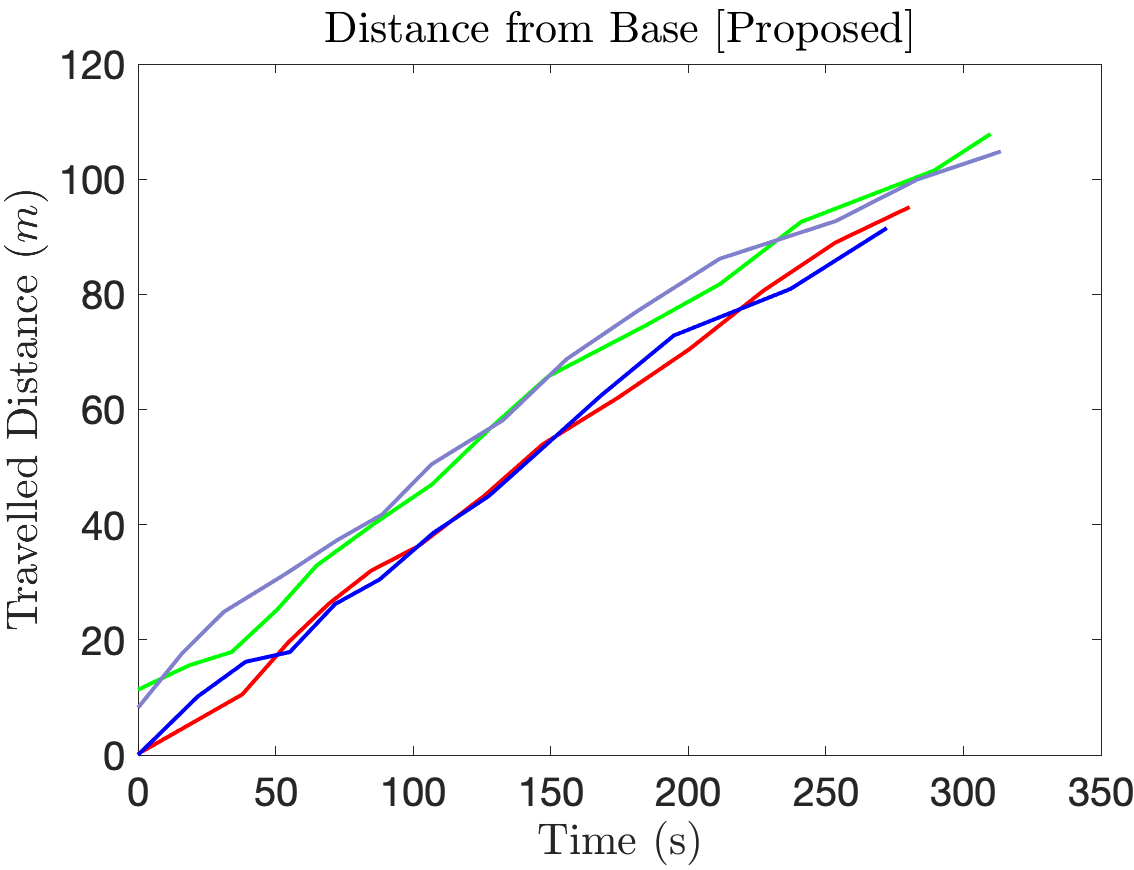}
        
    }
    \subfigure[]
    {
        \includegraphics[width=0.47\linewidth]{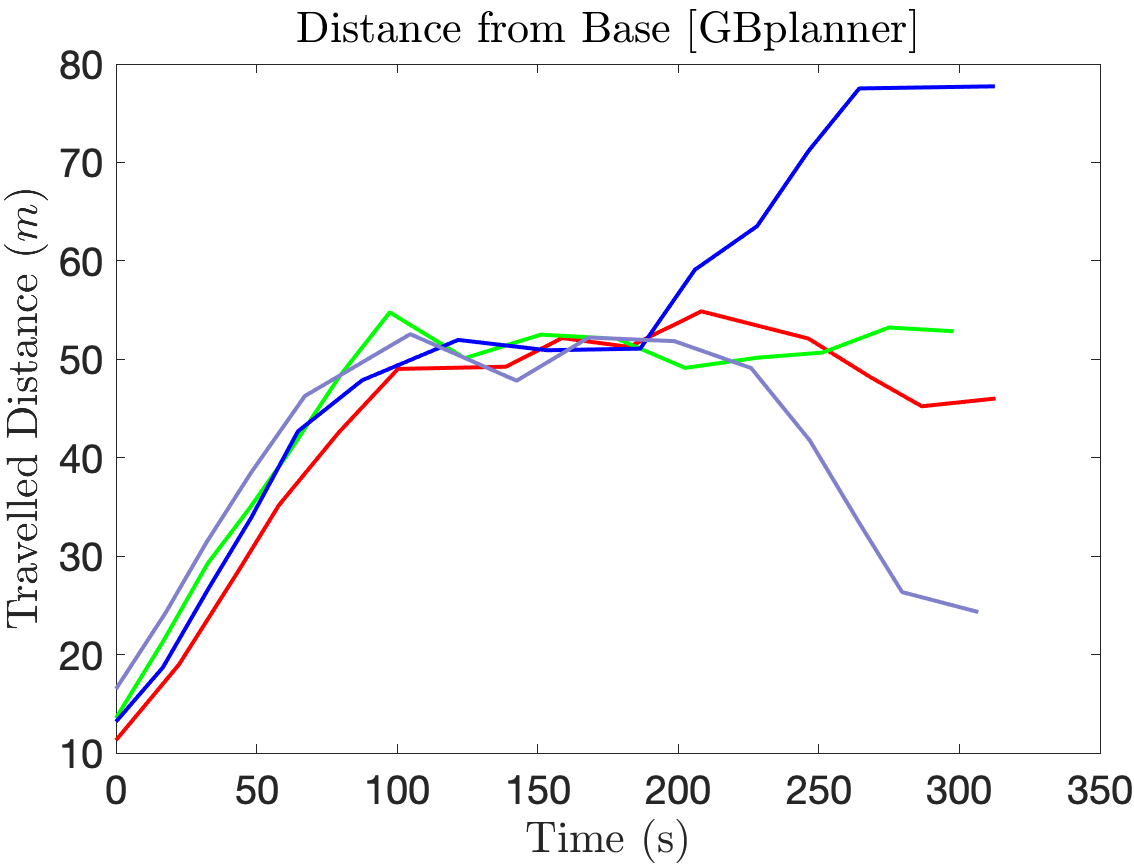}
        
    }
    \caption{Travelled Distance from base during exploration for 4 runs (a) Proposed approach (b) GB planner.}
    \label{fig:distancetravellednew}
\end{figure}

We also compute the volume of explored lava tube by our methods as well as graph based planner. As it is presented in \autoref{fig:volumeexplored}, the exploration volume is increased as the MCQ traverses through more unexplored areas in given time. The proposed framework is developed with the focus of efficiently utilizing the resource constrained MCQ's energy while in flight by rapidly traversing through unexplored areas. The expected behaviour is evident in \autoref{fig:volumeexplored} as using the proposed method the MCQ explores more volume of the environment. { In~\autoref{fig:volumeexplorednew} exploration metrics were derived based on the explored volume by the two methods. Multiple simulation runs in different environment are evaluated to make comparison show the higher exploration gain by proposed autonomy framework compared to the graph based exploration planner. In given fixed time budget based missions the proposed framework equipped MCQ also covers more ground as it is evident in \autoref{fig:distancetravellednew}. Moreover, the maximum velocity for exploration corresponding to~\autoref{fig:volumeexplored} is 0.8 $m/s$ and the maximum allowed velocity corresponding to the exploration metrics in~\autoref{fig:volumeexplorednew} are 1.5 $m/s$. Which also evidently explains the change in explored volume over time. In the distance metrics, the distance is calculated from the base. The decreasing distance for gbplanner in \autoref{fig:distancetravellednew} correspond to looping behaviour of the sampling based approach which further disregard the idea of efficiently utilizing available energy to maximize exploration. In contrast, as presented in \autoref{sec:IFE} the proposed approach is based on rapidly selecting future way points and planning paths to such points on the go in order to avoid hovering at same place. The effects on this on exploration volume gain and covered ground is evident in \autoref{fig:volumeexplorednew} and \autoref{fig:distancetravellednew}.} The graph based planner also performs very well in terms of exploration volume in given time due to the ability to re locate the MCQ to global frontier of interest when low local information gain is encountered. Nearly same exploration volume gain observed in the last 150 to 200 seconds because once the maximum exploration mission reaches either 10 or 15 minutes, homing is triggered in both cases to bring the MCQ back to start position. {In~\autoref{fig:fullmissioncompare} we have also considered an exploration mission in unconstrained time budged based scenarios to see the evaluate energy efficient exploration by the two framework. It is evident that even in longer missions, the proposed autonomy architecture enabled MCQ shows higher exploration gain as well as covers more distance from the base as shown in~\autoref{fig:fullmissioncompare}. The simulations were carried out with same configurations multiple times to validate the redundancy of the proposed system and the explored parts of lava tube in four different runs is presented in \autoref{fig:fourruns}.}

\begin{figure}[h!]
    \centering
    \subfigure[]
    {
        \includegraphics[width=0.47\linewidth]{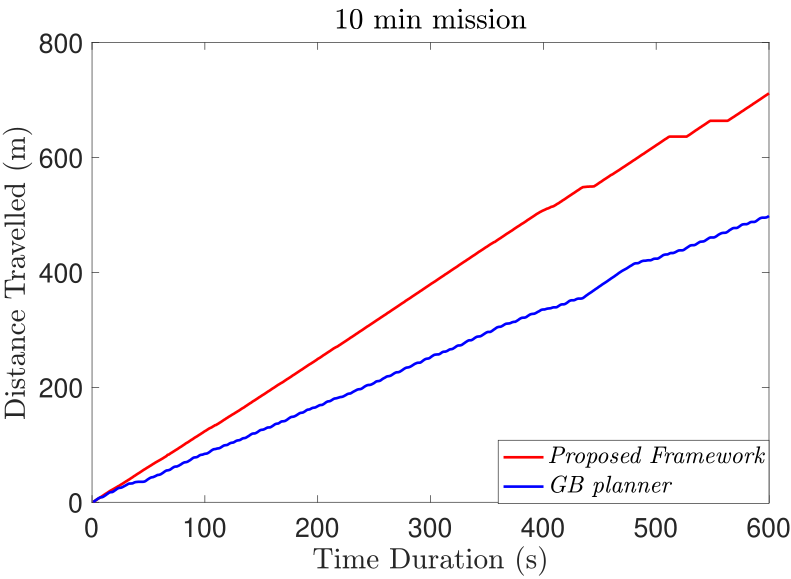}
        
    }
    \subfigure[]
    {
        \includegraphics[width=0.47\linewidth]{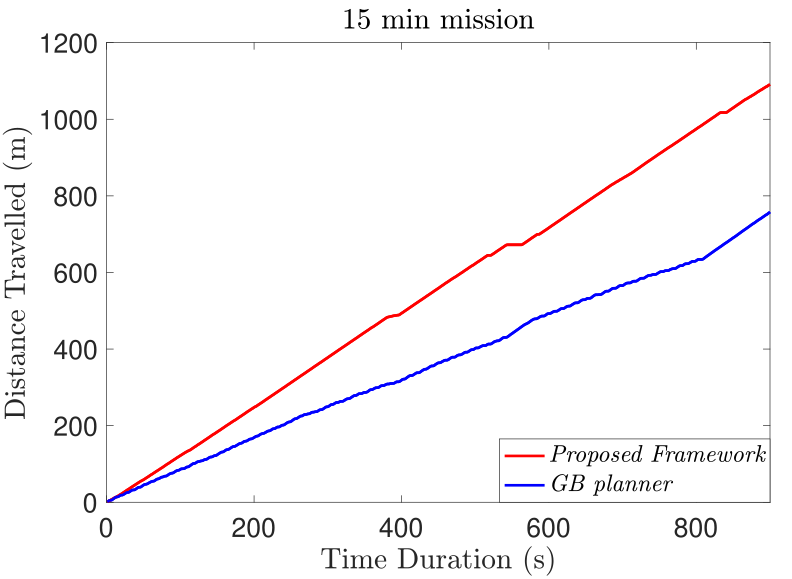}
        
    }
    \caption{Travelled Distance from the base during exploration (a) 10 minute mission (b) 15 minute mission.}
    \label{fig:distancetravelled}
\end{figure}

\begin{figure}[h!]
    \centering
    \subfigure[]
    {
        \includegraphics[width=0.47\linewidth]{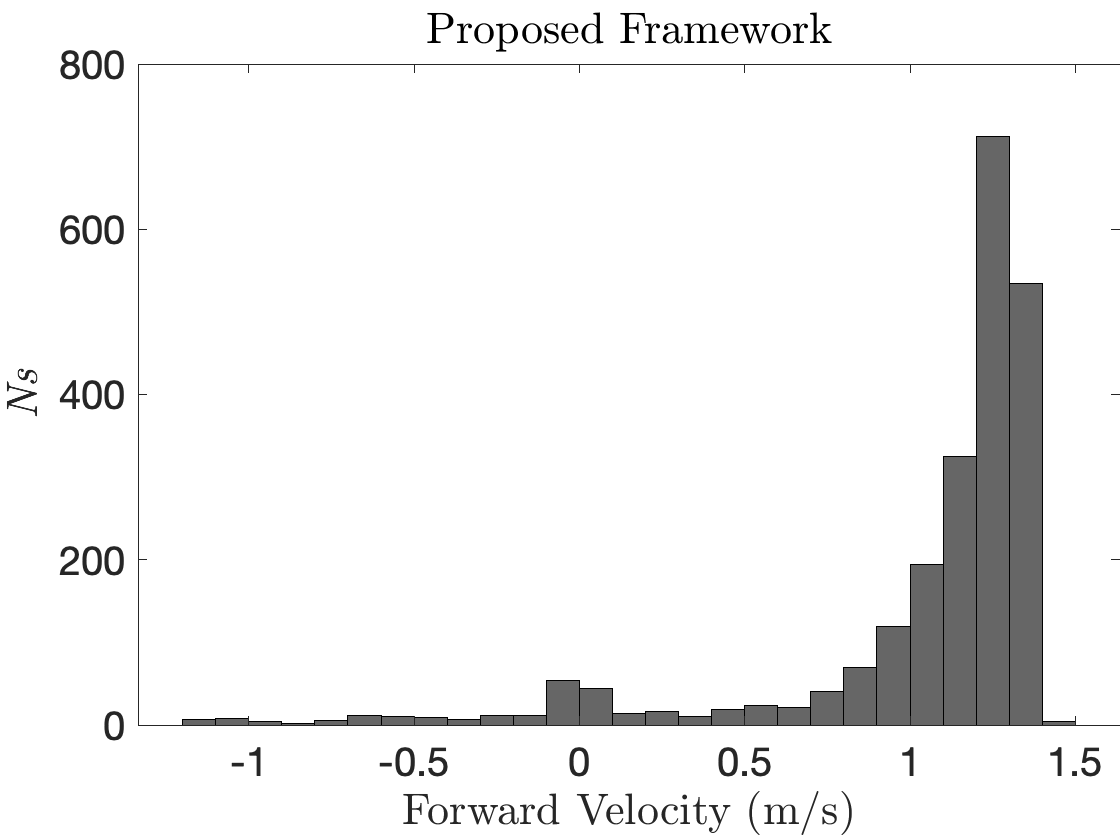}
        
    }
    \subfigure[]
    {
        \includegraphics[width=0.47\linewidth]{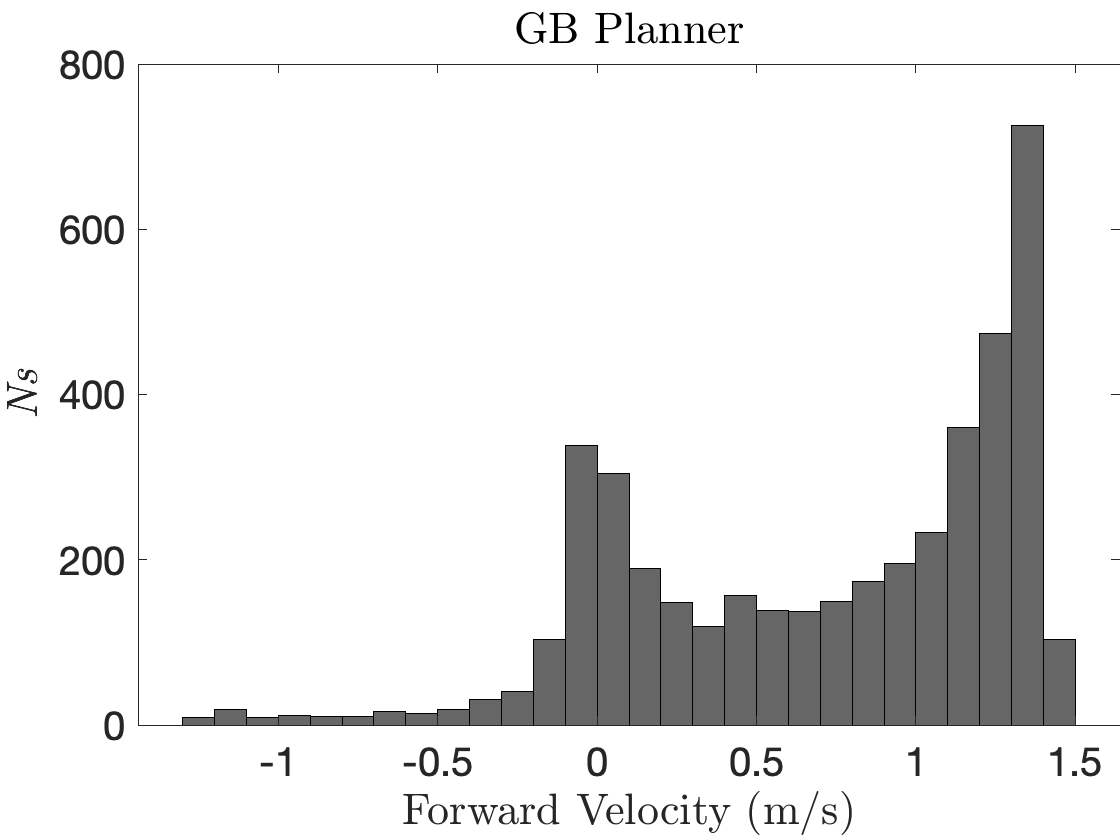}
        
    }
    {\caption{(a) Comparisons based on Samples (Ns) corresponding to rapid exploration behaviour at continuous forward velocity with minimal hovering.}}
    \label{fig:moveforwardvelocity}
\end{figure}

\begin{figure}[h!]
    \centering
    \subfigure[]
    {
        \includegraphics[width=0.47\linewidth]{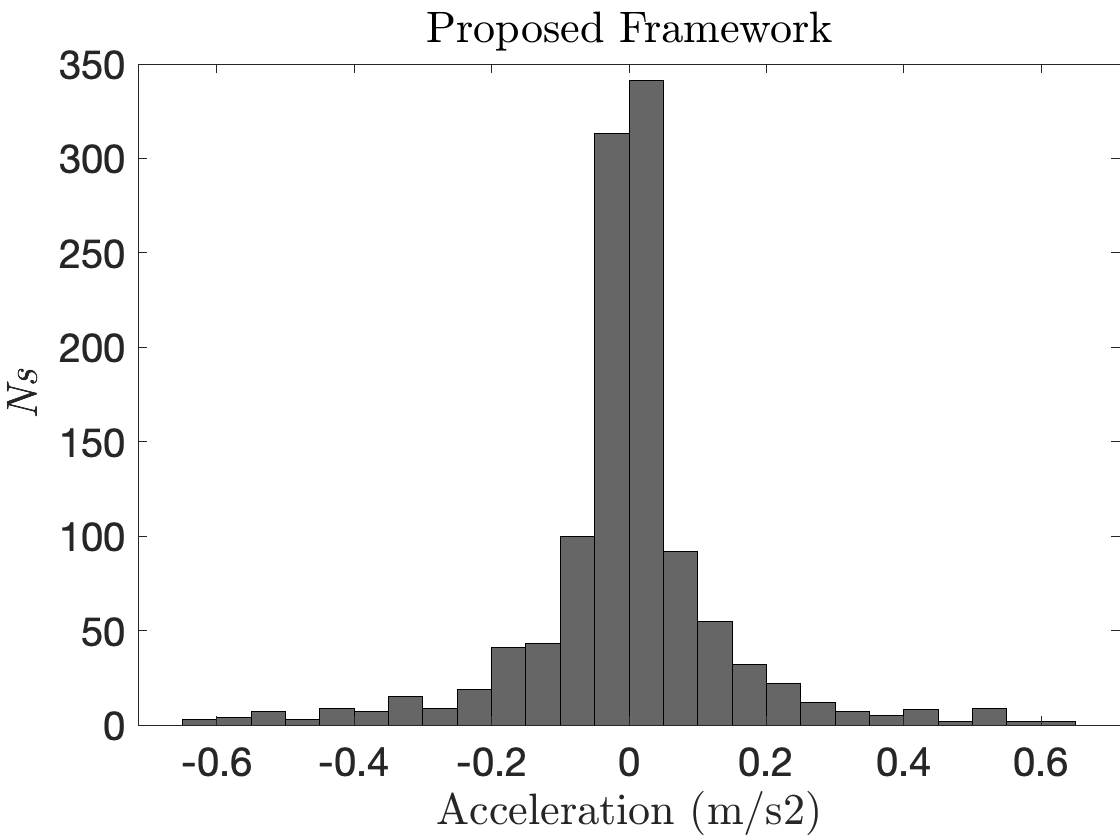}
        
    }
    \subfigure[]
    {
        \includegraphics[width=0.47\linewidth]{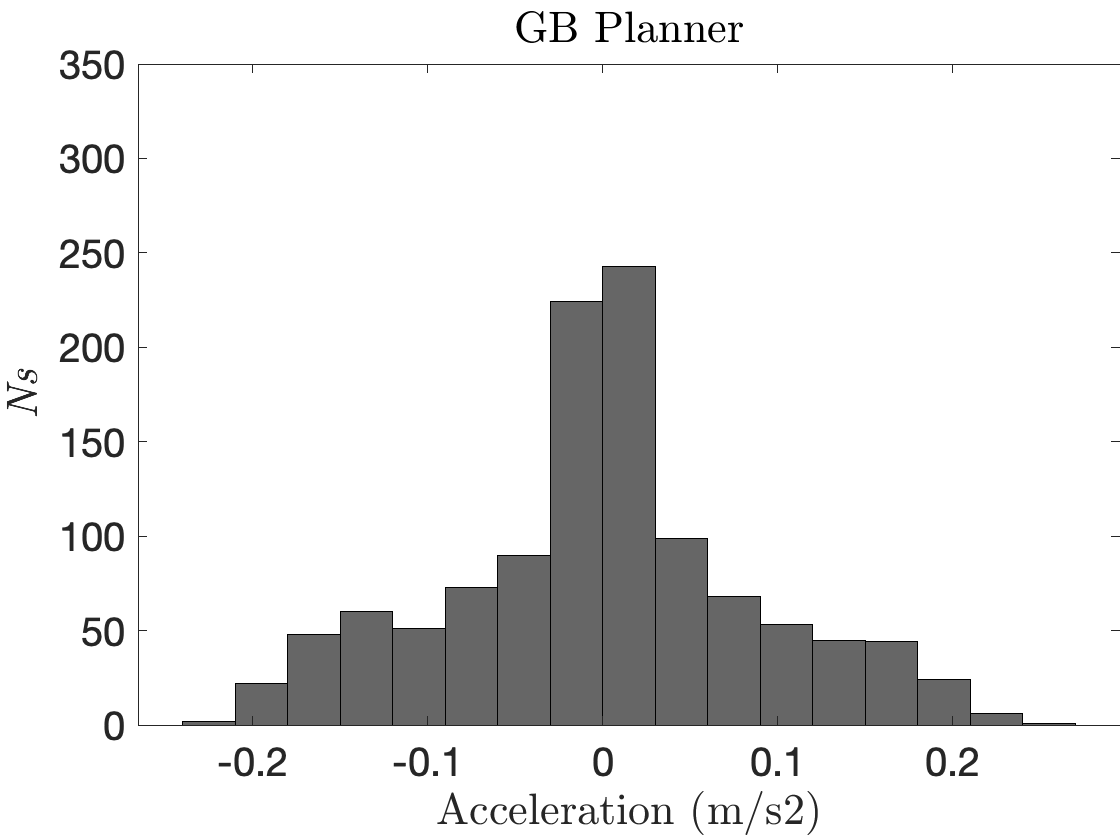}
        
    }
    {\caption{(a) Comparisons based on Samples (Ns) corresponding to change in forward velocity over $\delta T$.}}
    \label{fig:acceleration}
\end{figure}

\begin{figure}[h!]
    \centering
    \subfigure[]
    {
        \includegraphics[width=0.47\linewidth]{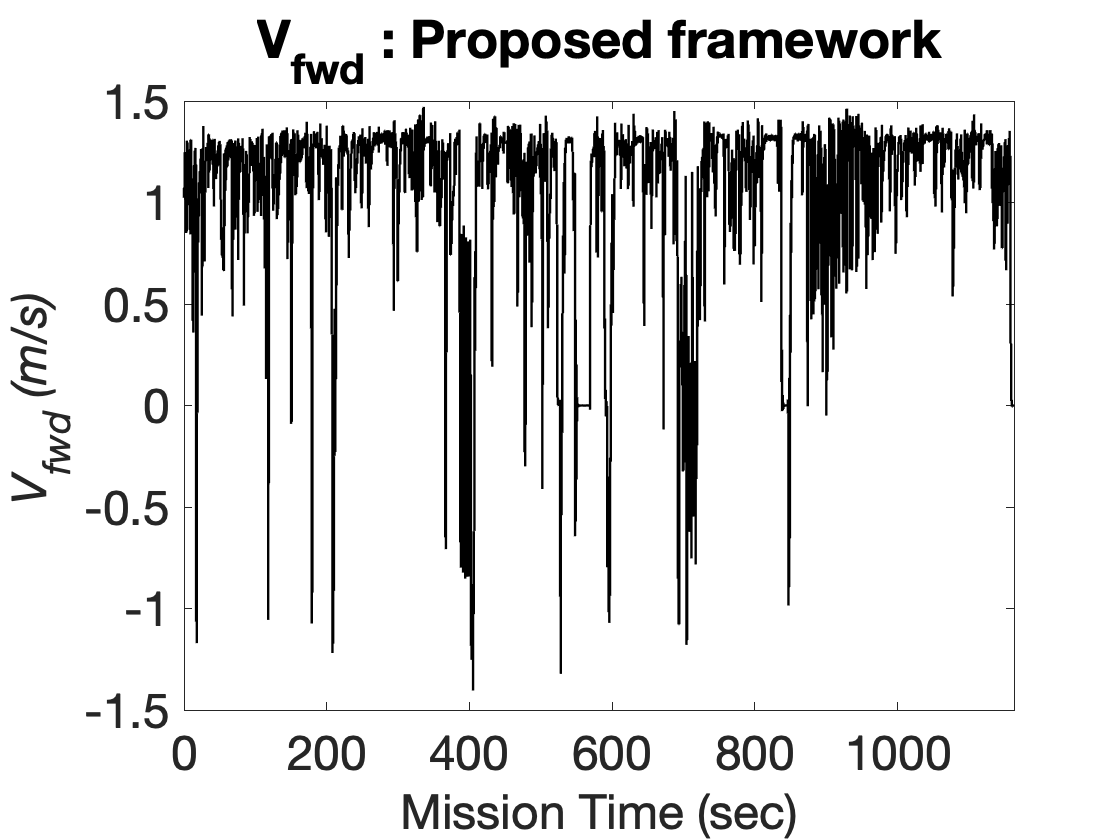}
        
    }
    \subfigure[]
    {
        \includegraphics[width=0.47\linewidth]{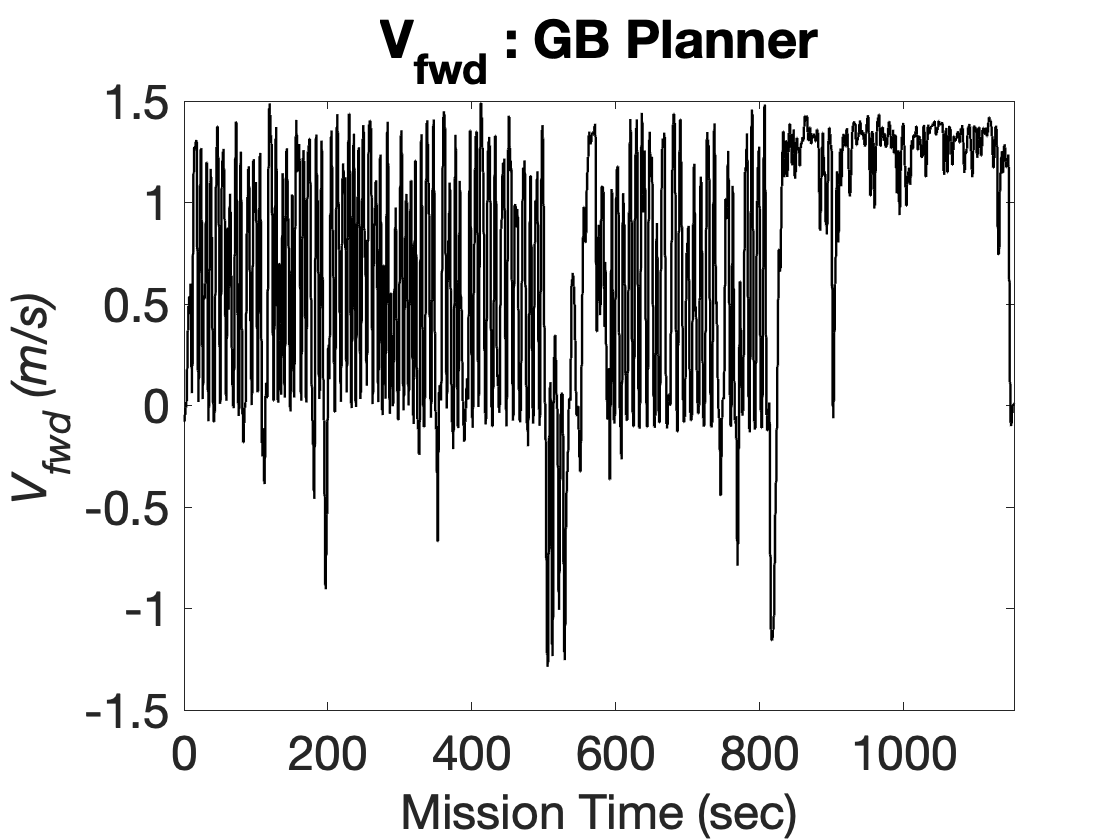}
        
    }
    \caption{(a) Forward Velocity of the MCQ (Proposed framework) (b) Forward Velocity of the MCQ (GB Planner).}
    \label{fig:forwardvelocities}
\end{figure}

\begin{figure}[h!]
    \centering
    \subfigure[]
    {
        \includegraphics[width=0.47\linewidth]{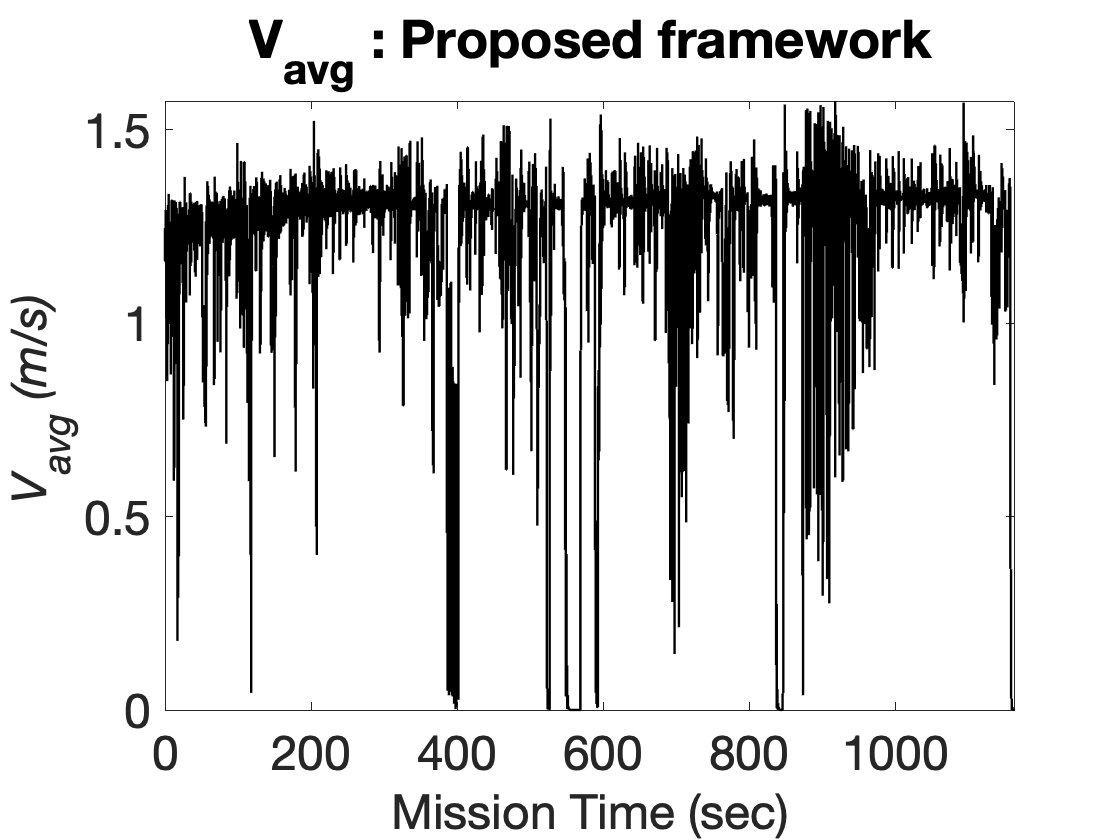}
        
    }
    \subfigure[]
    {
        \includegraphics[width=0.47\linewidth]{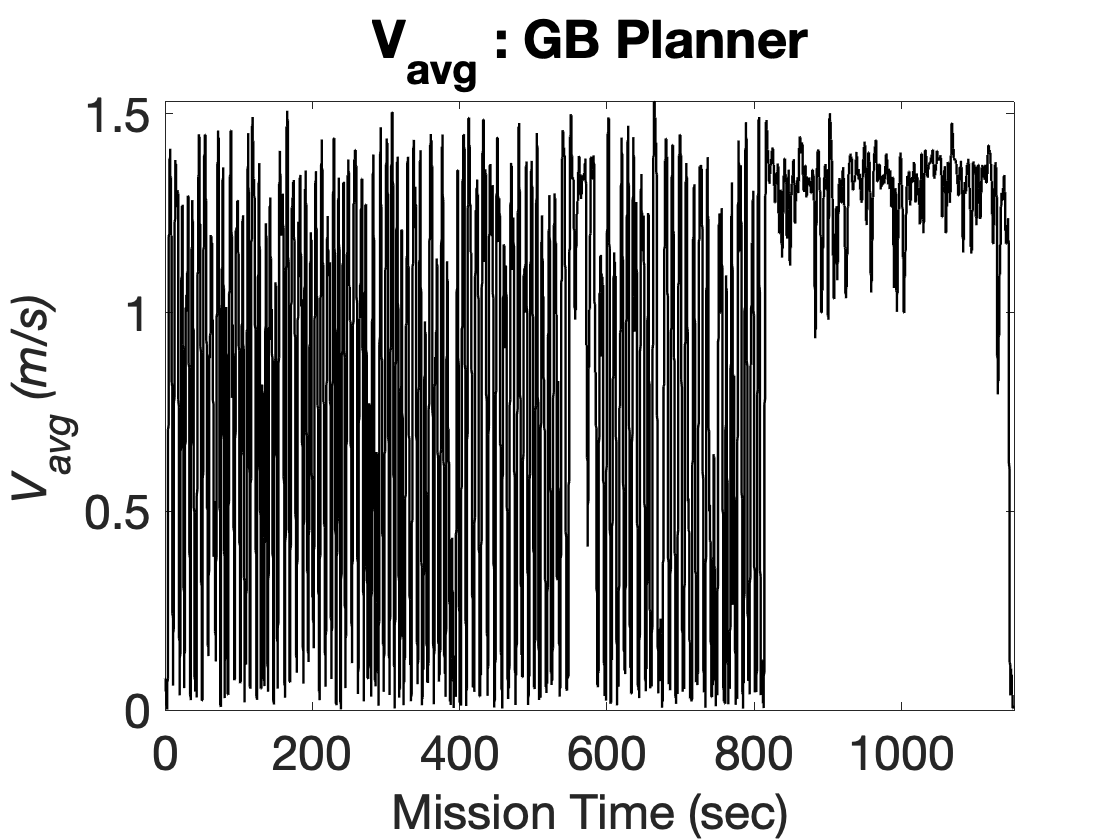}
        
    }
    \caption{(a) Average Velocity of the MCQ (Proposed framework) (b) Average Velocity of the MCQ (GB Planner).}
    \label{fig:avgvelocities}
\end{figure}

During the autonomous homing the MCQ travels back using the path computed in map which was built while exploring thus, not leading to more new unexplored places. We also compute the distance travelled by the MCQ in both methods in a 10 minute and 15 minute exploration missions. As described earlier, the proposed framework is designed to rapidly explore Martian lava tube in order to maximize the ground covered in given time. As shown in \autoref{fig:distancetravelled} the proposed framework allows the MCQ to cover more ground in a time constraint based exploration scenario as compared to the graph based planner. The graph based planner also shows a nearly linear increase in distance travelled by the MCQ. However, the difference in covered distance in comparison is mainly due to the $Plan \rightarrow\ Execute \rightarrow\ Stop \rightarrow\ Plan \rightarrow\ Execute...$ nature of the graph based planner. Since the gbplanner is designed and tested intensively in field experiments, such planning nature has turned out to be considerably safe for MAVs in subterranean exploration scenarios. For fair comparison we also keep the maximum velocity of the MCQ to be exactly the same ($1.5 m/s$) as it plays big role in exploration volume gain. In order to efficiently use the available energy of the MCQ it is important to maximize the forward flight velocity while exploring. In order to demonstrate rapid exploration behaviour the forward velocities of MCQ are plotted against the simulation mission time using the data acquired from recorded rosbags using both methods in 15 minute exploration. In \autoref{fig:forwardvelocities} and \autoref{fig:avgvelocities} it is evident that using the proposed method the MCQ aims at maintaining similar forward velocity in order to efficiently cover more ground using available energy. The state of the art method uses the RRT based nodes to compute exploration gain thus the MCQ hovers at same place in between planning iterations. The result of which is seen as high fluctuation in forward velocity as presented in \autoref{fig:forwardvelocities}. Moreover such behaviour is expected given that the planner iteratively plans short yet safe segmented paths based on local information gain as mentioned in~\citep*{dang2020graph}. The negative values of forward velocity simply indicate that the MCQ pitched backward to travel in reverse direction with minimal change in heading. As discussed previously, the MAV's energy is consumed optimally by continuously flying forward at high speeds such that more ground can be covered in given time. We demonstrate this by comparing the accelerations samples and velocity samples during the exploration flight. \autoref{fig:moveforwardvelocity} shows the velocity sampled during exploration. From \autoref{fig:moveforwardvelocity} it is evident that the proposed framework shows more number of samples which correspond to high forward velocities as compared to the graph based planner. The proposed method also shows low samples which correspond to approximately zero forward velocity (near hover point in flight) due to the \textit{compute-frontier-plan-path} while \textit{exploring} nature of exploration of the proposed approach. In addition to that \autoref{fig:acceleration} shows the acceleration samples for both methods. From \autoref{fig:acceleration} is is clear that using proposed approach the MCQ continuously flies forward as it explores exhibiting higher samples that correspond to nearly zero accelerations. This clearly shows the exploration performance of the proposed method in terms of efficiently utilizing the energy of MCQ to explore the large lava tube environment on Mars. 

\begin{figure}[h!]
  \centering
    \includegraphics[width=\linewidth]{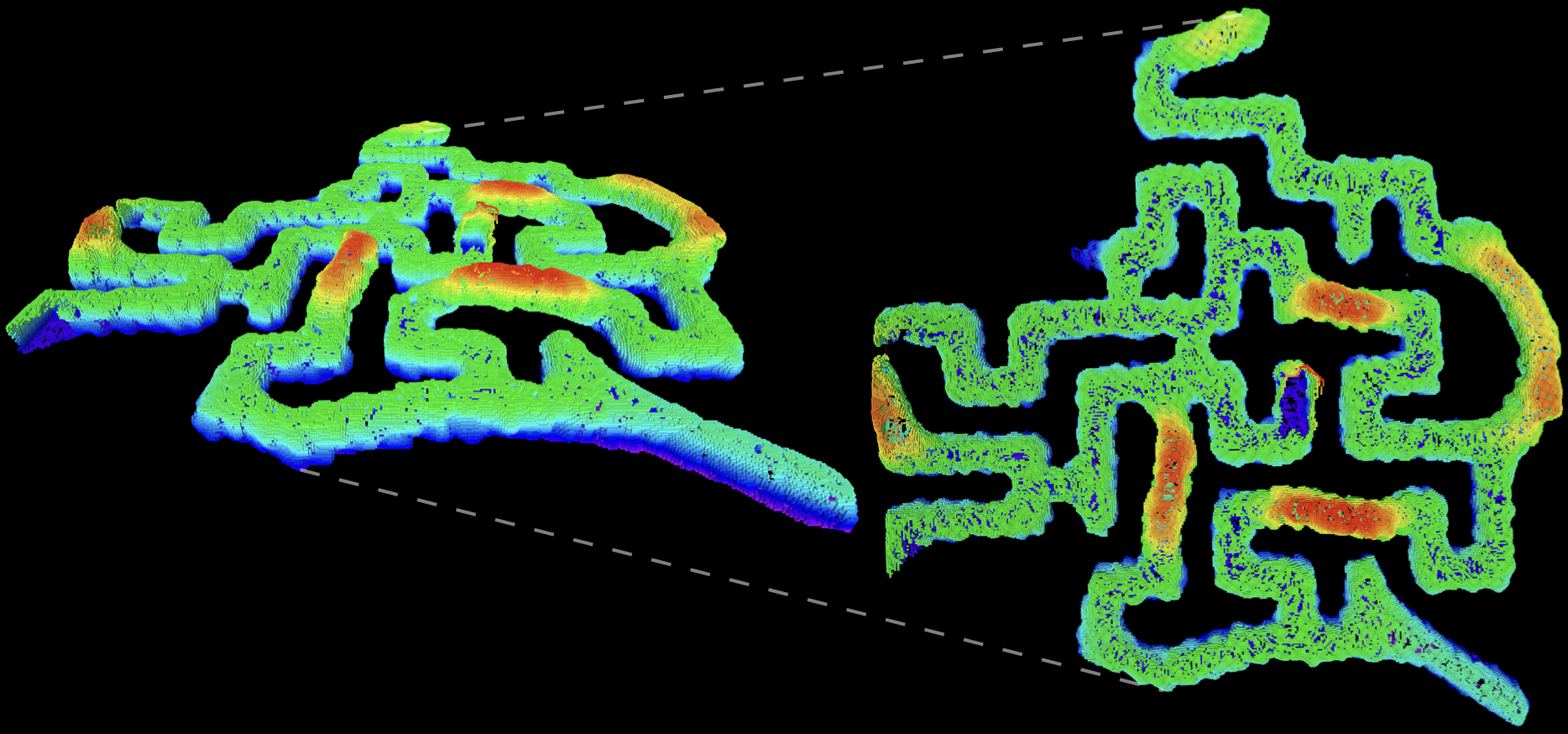}
    {\caption{nstructed occupancy based map of the explored area by the proposed autonomy architecture on MCQ.
    }}
  \label{fig:mapbuilt}
\end{figure}

In order to show the nearly complete (99$\%$) exploration of the Mars lava tube using the proposed method, the Octomap built during the exploration mission is presented in \autoref{fig:mapbuilt}. The nearly complete exploration mission is achieved in ~1700 seconds using the proposed approach.

\begin{figure}[h!]
  \centering
    \includegraphics[width=0.8\linewidth, angle=180]{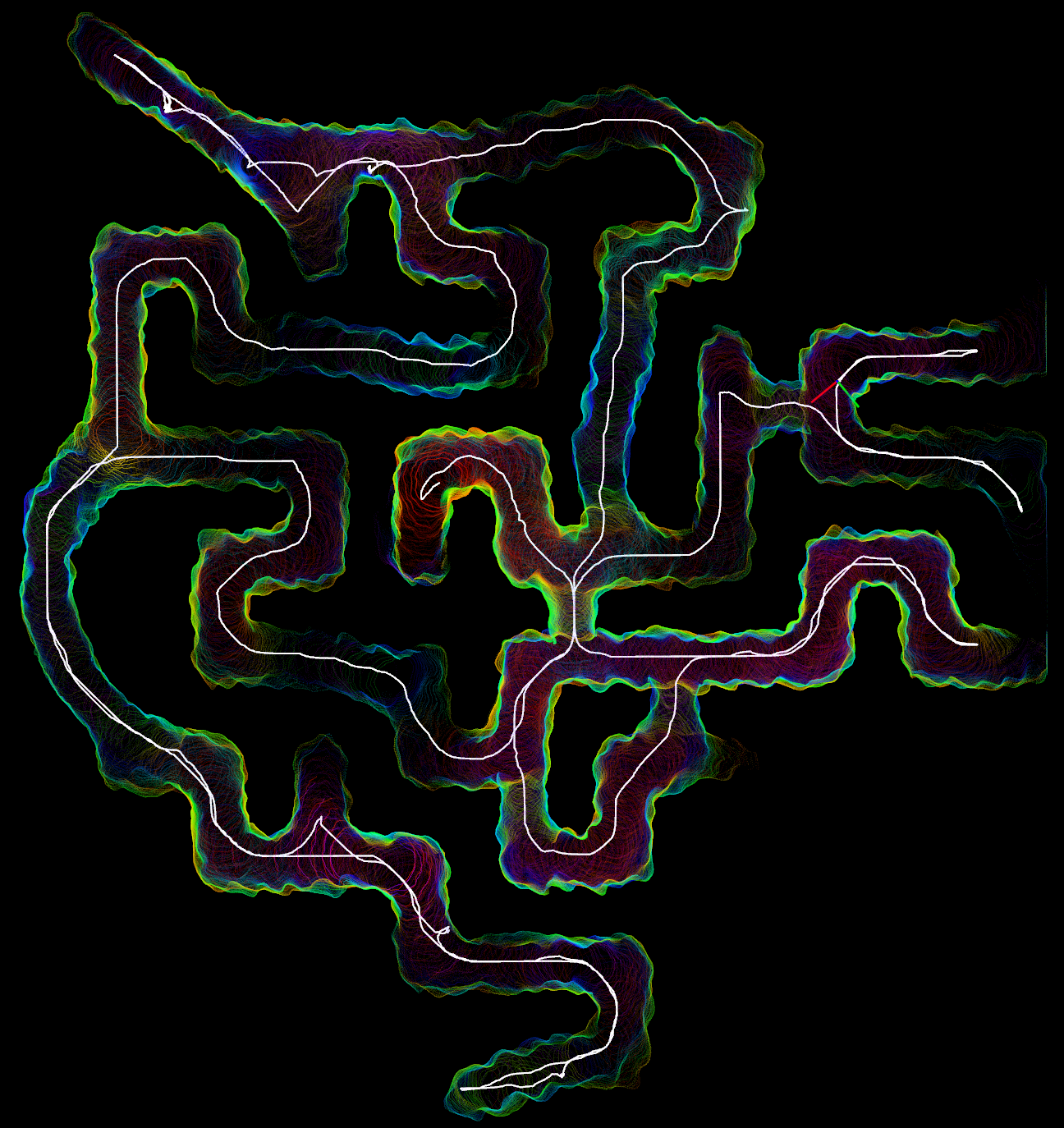}
    \caption{Point cloud map and MCQ's tracked trajectory after 99$\%$ exploration of Mars lava tube environment using the presented method.}
  \label{fig:pclmap}
\end{figure}

\section{Concluding Remarks and Future Directions}\label{sec:conclusions}

{Subsurface voids and lava tubes are prime candidates for search of life on Mars. Such sites are also areas of interest for potential habitats for future colonization of Mars. Robotic exploration and surveying of planetary bodies further strengthen our understandings of terrain in order to prepare humanity for future Mars missions. The Mars helicopter surpassed the expectations by proving the first powered flight in Martian thin atmosphere. The Mars helicopter also proved to be an excellent candidate to be aerial scout for the ground rovers on Mars in order to assist in planning safe paths. However, in regard to the application of exploring subsurface voids and lava tube channels on Mars, the technology demonstrator vehicle lacks autonomy in certain aspects of independent decision making while exploring. In response, inspired from the Ingenuity helicopter, this work focuses on the application of energy efficient exploration of subterranean areas on Mars. The importance of efficiently utilizing the energy from batteries to make the most out of the flight is huge, considering the thin atmosphere and low surface pressure on Mars. This work proposed an energy preserving incremental frontiers based exploration approach coupled with risk-aware planning and advanced control scheme to complement overall autonomy architecture of the Mars Coaxial Quadrotor (MCQ). The proposed exploration approach is formulated with the goal of rapidly exploring forward while minimizing motor saturating yaw movements. The proposed approach considers directly accessible frontiers ahead of the MCQ to prioritize forward exploration and also consist of a dedicated cost based global re positioning scheme in order to navigate the MCQ to partially seen area in case of dead end in exploration. The proposed scheme was further validated through extensive simulations to evaluate the energy efficient exploration and redundancy of the autonomy architecture. We also compare our method with state of the art graph based exploration planner (designed for subterranean exploration) in order to show the exploration gain and covered ground in fixed time budged based exploration missions. The comparisons clearly show the shortcomings of sampling based approach when energy efficient exploration is targeted due to the lack of focused exploration goals generation in one directions. Through comparisons, the proposed exploration method also showed the impact on increased exploration gain by computing future exploration goals and plan paths on the go. In conclusion, we propose the autonomy architecture for a potential future Mars Coaxial Quadrotor that considers energy efficient exploration with redundant systems and completely autonomous operations in which mission plan could be loaded from Earth and the MCQ autonomously executes exploration mission on Mars and returns with accurate 3D map of the lava tube or subsurface void.} 

{The design of the MCQ was proposed in our previous work which is still at early stage in design for a space craft class vehicle to operate on Mars. Therefore as part of future direction, the Design of vehicle to operate in thin Martian atmosphere could be further improved with focus on increasing payload size to equip more sophesticated sensor suite. Moreover, experimental analysis of the vehicle to operate in controlled atmospheric chambers could be performed to further optimize the design. The planned future directions also include evaluating modern control frameworks by dropping the vehicle from stratospheric balloons to evaluate hover and trajectory tracking of Mars coaxial quadrotor at high altitudes that somewhat resemble actual flight conditions on Mars. On the navigation side, the autonomy components could be further evaluated in presence of external disturbances such as wind gusts etc. The idea of swarms of Mars coaxial quadrotors but with smaller size vehicles to perform multiple autonomy tasks to assist the rovers and other robots in high level decision making process is one of the challenging yet highly rewarding research direction that could benefit from the proposed autonomy architecture.}




\section{Acknowledgement}

This work has been partially funded by the European Unions Horizon 2020 Research and Innovation Programme under the Grant Agreement No. 869379 illuMINEation.


\bibliographystyle{jasr-model5-names}
\biboptions{authoryear}
\bibliography{mybib}

\end{document}